\documentclass[letterpaper]{llncs}
\usepackage[T1]{fontenc}
\usepackage{graphicx}
\usepackage{cite}
\usepackage{xcolor}
\usepackage{adjustbox}
\usepackage{multirow}
\usepackage{caption}
\usepackage{subcaption}
\usepackage{amsmath}
\usepackage[ruled, vlined]{algorithm2e} 
\usepackage{makecell} 
\usepackage{lipsum}
\usepackage{setspace}

\makeatletter

\begin{document}
\title{ST-CoNAL: Consistency-Based Acquisition Criterion Using Temporal Self-Ensemble for Active Learning}
\titlerunning{ST-CoNAL: Consistency-Based Active Learning}
\author{Jae Soon Baik \and
In Young Yoon \and Jun Won Choi\thanks{Corresponding Author}}
\authorrunning{J.S. Baik et al.}
\institute{Hanyang University, Seoul, Korea\\
\email{\{jsbaik, inyoungyoon\}@spa.hanyang.ac.kr,}  \email{junwchoi@hanyang.ac.kr}}
\maketitle              % typeset the header of the contribution
\begin{abstract}
Modern deep learning has achieved great success in various fields. However, it requires the labeling of huge amounts of data, which is expensive and labor-intensive. Active learning (AL), which identifies the most informative samples to be labeled, is becoming increasingly important to maximize the efficiency of the training process. The existing AL methods mostly use only a single final fixed model for acquiring the samples to be labeled. This strategy may not be good enough in that the structural uncertainty of a model for given training data is not considered to acquire the samples.  In this study, we propose a novel acquisition criterion based on temporal self-ensemble generated by conventional stochastic gradient descent (SGD) optimization. These self-ensemble models are obtained by capturing the intermediate network weights obtained through SGD iterations.   Our acquisition function relies on a {\it consistency measure} between the student and teacher models. The student models are given a fixed number of temporal self-ensemble models, and the teacher model is constructed by averaging the weights of the student models. Using the proposed acquisition criterion, we present an AL algorithm, namely {\it student-teacher consistency-based AL} (ST-CoNAL). Experiments conducted for image classification tasks on CIFAR-10, CIFAR-100, Caltech-256, and Tiny ImageNet datasets demonstrate that the proposed ST-CoNAL achieves significantly better performance than the existing acquisition methods. Furthermore, extensive experiments show the robustness and effectiveness of our methods.

\keywords{Active Learning \and Consistency  \and Temporal Self-Ensemble \and Image Classification .}
\end{abstract}

\section{Introduction}

Deep neural networks (DNNs) require a large amount of  training data to optimize millions of weights. In particular, for supervised-learning tasks, labeling of training data by human annotators is expensive and time-consuming. The labeling cost can be a major concern for machine learning applications, which requires a collection of real-world data on a massive scale (e.g., autonomous driving) or the knowledge of highly trained experts for annotation (e.g., automatic medical diagnosis).  Active learning (AL) is a promising machine learning framework that maximizes the efficiency of labeling tasks within a fixed labeling budget \cite{settles2009activesurvey}.

This study focuses on the pool-based AL problem, where the data instances to be labeled are selected from a pool of unlabeled data. In a pool-based AL method, the decision to label a data instance is based on a {\it sample acquisition function}. The acquisition function, $a(x,f)$, takes the input instance $x$ and the currently trained model $f$ and produces a score to decide if $x$ should be labeled. Till date, various types of AL methods have been proposed \cite{shannon1948mathematical, gal2016dropout, gal2017dbal, settles2009activesurvey, yoo2019learning, sinha2019variational, zhang2020state, kim2021task, gao2020consistency}. The predictive uncertainty-based methods \cite{shannon1948mathematical, gal2016dropout, gal2017dbal, yoo2019learning} used well-studied theoretic measures such as entropy and mutual information. % that are based on model prediction. 
Recently, representation-based methods \cite{sinha2019variational, sener2017coreset, kim2021task, zhang2020state} have been widely used as a promising AL approach to exploit high-quality representation of DNNs. However, the acquisition used for these methods rely on a single trained model $f$, failing to account for model uncertainty arising given a limited labeled dataset. To solve this problem, several AL methods \cite{beluch2018thepower, settles2009activesurvey} attempted to utilize an ensemble of DNNs and design acquisition functions based on them. 
However, these methods require significant computational costs to train multiple networks.

When the amount of labeled data is limited, semi-supervised learning (SSL) is another promising machine learning approach to improve performance with low labeling costs. SSL improves the model performance by leveraging  a large number of unlabeled examples \cite{oliver2018realistic}.  {\it Consistency regularization} is one of the most successful approach to SSL \cite{gao2020consistency, laine2016temporal,miyato2018virtual,tarvainen2017meanteacher}. In a typical semi-supervised learning, the model is trained using the consistency-regularized loss function  $\mathcal{E}_{\rm ce}+\lambda \mathcal{E}_{\rm con}$, where $\mathcal{E}_{\rm ce}$ denotes the cross-entropy loss and $\mathcal{E}_{\rm con}$ denotes the consistency-regularized loss.  Minimization of $\mathcal{E}_{\rm con}$ regularizes the model to produce consistent predictions over the training process, improving the performance for a given task. % In \cite{laine2016temporal, Sajjadi2016Regularization},  $\mathcal{E}_{\rm con}$ relies on the predictions produced for perturbed inputs or models. 
%The most difficult thing when using consistency regularization is that the model produces unstable prediction for unlabeled sample, which leads to inaccurate consistency loss.
%Various methods have been proposed to measure the consistency of predictions from DNNs \cite{laine2016temporal, tarvainen2017meanteacher, oliver2018realistic, miyato2018virtual}. 
$\Pi$ model \cite{laine2016temporal} applied a random perturbation to the input of the DNN and measured the consistency between the model outputs. Mean Teacher (MT) \cite{tarvainen2017meanteacher} produced the temporal self-ensemble through SGD iterations and measured the consistency between the model being trained and the teacher model obtained by taking the exponential moving average (EMA) of the self-ensemble. These methods successfully  regularized the model to produce consistent predictions while using perturbed predictions on unlabeled samples.

\begin{figure*}[t]
\begin{center}
\fbox{\includegraphics[width=0.90\textwidth]{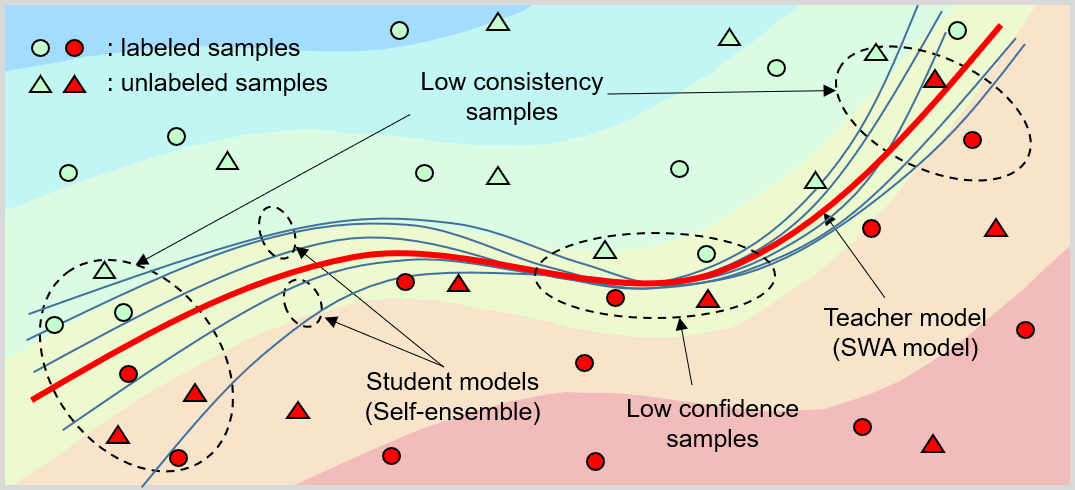}}
\end{center}
   \caption{{\bf Acquisition criterion}: Both labeled and unlabeled samples are represented in the feature space in the binary classification problem. {\it Low consistency samples} produce the predictions which significantly differ between the student and teacher models. {\it Low confidence samples} are found near the decision boundaries of the student models. The proposed ST-CoNAL evaluates the consistency measure for each data sample and selects those with the highest consistency score.}
\label{fig:decision_boundary}
\end{figure*}

The objective of this study is to improve the sample acquisition criterion for pool-based AL methods. Inspired by the consistency regularization for SSL, we build a new sample acquisition criterion that measures consistency between multiple ensemble models obtained during the training phase.  The proposed acquisition function generates the temporal self-ensemble by sampling the models at the intermediate check-points of the weight trajectory formed by the SGD optimization. This provides a better acquisition performance and eliminates the additional computational cost required by previous ensemble-based AL methods. In \cite{athiwaratkun2018fastswa, izmailov2018averagingswa, loshchilov2017sgdr}, the aforementioned method has shown to produce good and diverse self-ensembles, which were used to improve the inference model via {\it stochastic weight averaging} (SWA) \cite{athiwaratkun2018fastswa, izmailov2018averagingswa, loshchilov2017sgdr}. We derive the acquisition criterion based on the temporal self-ensemble models used in SWA. We present the AL method, referred to as {\it student-teacher consistency-based AL (ST-CoNAL)}, which measures the consistency between the student and teacher models. The self-ensemble model constitutes a student model, and  a teacher model is formed by taking an equally-weighted average (EWA) of the parameters of the student models. Treating the output of the teacher model as a desired supervisory signal, ST-CoNAL measures the Kullback–Leibler (KL) divergence of each teacher-student output pairs. The acquisition function of ST-CoNAL acquires the samples to be labeled that yield the highest inconsistency. 

Though ST-CoNAL was inspired by the consistency regularization between student and teacher models of the MT, these two methods are quite different in the following aspects. MT constructs the teacher model by assigning larger weights to more recent model weights obtained through SGD iterations. Due to this constraint, as training progresses, the teacher model in MT tends to be correlated with the student models, making it difficult to measure good enough consistency measure for AL acquisition. To address this problem, ST-CoNAL generates a better teacher model by taking an equally-weighted averaging (EWA) of the weights of the student models instead of EMA used in MT. Similar to previous AL methods \cite{gal2016dropout, gal2017dbal} that utilize ensemble models to capture the posterior distribution of model weights for a given data set, the use of student model weights allows our acquisition criterion to account for model uncertainty.

We further improve our ST-CoNAL method by adopting the principle of entropy minimization used for SSL \cite{grandvalet2005semient, lee2013pseudo, sohn2020fixmatch, berthelot2019remixmatch, berthelot2019mixmatch, xie2019UDA}. We apply one of the entropy minimization methods, sharpening to the output of the teacher model. When the sharpened output is used for our KL divergence, our acquisition criterion can measure the uncertainty of the prediction for the given sample. Our evaluation shows that the ST-CoNAL is superior to other AL methods on various image classification benchmark datasets.

Fig.\ref{fig:decision_boundary} illustrates that these low consistency samples lie in the region of the feature space where the student models produce the predictions of a larger variation. Note that these samples are not necessarily identical to the low confidence samples located near the decision boundary specified by the teacher model. The proposed ST-CoNAL prefers the acquisition of the inconsistent samples rather than the low-confidence samples. 

The main contributions of our study are summarized as follows:
\begin{itemize}
    \item We propose a new acquisition criterion based on the temporal self-ensemble. Temporal self-ensemble models are generated by sampling DNN weights through SGD optimization.  ST-CoNAL measures the consistency between these self-ensemble models and acquires the most inconsistent samples for labeling. We evaluated the performance of ST-CoNAL on four different public datasets for multi-class image classification tasks. We observe that the proposed ST-CoNAL method achieves the significant performance gains over other AL methods.

    \item We identified a work relevant to ours \cite{gao2020consistency}. While both ours and their work aim to exploit consistency regularization for AL, our work differs from theirs in the following aspects. While CSSAL \cite{gao2020consistency} relies on the input perturbation to a single fixed model, ST-CoNAL utilizes the self-ensemble models to measure the consistency. Note that the benefits of using model ensembles for AL have been demonstrated in \cite{beluch2018thepower} and our findings about the superior performance of ST-CoNAL over CSSAL are consistent with the results of these studies. 
   
\end{itemize}

\section{Related Work}
\label{Related Work}

\subsection{Acquisition for AL} % \emph{uncertainty-based}, 2) \emph{consistency-based} and 3) \emph{diversity-based methods}.
The acquisition methods for pool-based AL can be roughly categorized into three classes: 1) \emph{uncertainty-based}, 2) \emph{representation-based} and 3) \emph{consistency-based methods}. The uncertainty-based methods \cite{beluch2018thepower, gal2016dropout, gal2017dbal, joshi2009multi,  lewis1994leastconfi, tong2001supportvec} estimate the prediction uncertainty for the given data instances. Max-entropy \cite{lewis1994leastconfi, shannon1948mathematical}, least confidence \cite{lewis1994leastconfi} and variation ratio \cite{johnson1966freeman_variation} are the widely used criteria.  Recently, the representation-based method \cite{sinha2019variational, zhang2020state, kim2021task, sener2017coreset} significantly improved the acquisition performance. VAAL \cite{sinha2019variational} performed adversarial learning to find informative unlabeled data in the task-agnostic latent representation of the data. TA-VAAL \cite{kim2021task} is the latest state-of-the-art method, which incorporated a task-aware uncertainty-based approach to the VAAL baseline. The consistency-based methods measure the disagreement among the predictions of the ensemble models \cite{gal2016dropout, gal2017dbal, zoubin2011dbal, gao2020consistency}. TOD-Semi \cite{huang2022temporalTOD}, {\it Query-by-committee} \cite{settles2009activesurvey}, and DBAL \cite{gal2017dbal}  are the well known criteria used in these methods.

\subsection{Consistency Regularization for SSL}
One of the most successful approaches to SSL is \emph{consistency regularization}. The $\Pi$ model \cite{laine2016temporal, Sajjadi2016Regularization} enforces consistency for  model $f_{\theta}(x)$ by minimizing $d(f_{\theta}(x), f_{\theta}(\hat{x}))$, where $\hat{x}$ denotes the data perturbed by noise and augmentation, and $d(a,b)$ measures the distance between $a$ and $b$. While $\Pi$ model uses the predictions as a target, the consistency loss for regularizing the consistent behavior between an ensemble prediction and the model prediction was shown to be effective.
{\it Mean teacher} method \cite{tarvainen2017meanteacher} generates the weights of the teacher model with EMA of the student weights over SGD iterations. ICT \cite{verma2019ict} and MixMatch \cite{berthelot2019mixmatch} enforce consistency between linearly interpolated inputs and model predictions.
 
\subsection{Temporal Self-Ensemble via Stochastic Weight Averaging}
Several studies have attempted to use temporal self-ensemble obtained through model optimization to improve the inference performance \cite{laine2016temporal, chen2021temporalSSOD, lee2019robustLTEC, izmailov2018averagingswa}. Izmailov et al. \cite{izmailov2018averagingswa} proposed SWA method, which determines an improved inference model using the equally weighted average of the intermediate model weights traversed via SGD iterates. Using cyclical learning rate scheduling \cite{loshchilov2017sgdr}, SGD can yield the weights in the flat regions in the loss surface. 
By averaging the weights found at SGD trajectory, SWA can find the weight solution that can generalize well. 
In \cite{athiwaratkun2018fastswa}, consistency regularization and modified cyclical learning rate scheduling assisted in finding the improved weights through SWA. 

\subsection{Combining AL and SSL} The combination of SSL and AL was discussed in several works \cite{Thomas2019Active, gal2017dbal, gao2020consistency, Rhee2017Active, tomanek2009fusion, zhu2003combining}. However, most works considered applying SSL and AL independently. Only a few works considered the joint design of SSL and AL. Zhu et al. \cite{zhu2003combining}  used a Gaussian random field on a weighted graph to combine SSL and AL. Additionally, the authors of \cite{gao2020consistency} proposed an enhanced acquisition function by adopting the variance over the data augmentation of the input image.

\section{Proposed Active Learning}
In this section, we present the details of the ST-CoNAL algorithm. 

\subsection{Consistency Measure-Based Acquisition}
\label{ST-CoNAL acquisition function}
The proposed algorithm follows the setup of a pool-based AL for the $K$-image class classification task. Suppose that the labeling budget is fixed to $b$ data samples. In the $j$th sample acquisition step, we are given the dataset $\mathcal{D}^{j}$, which consists of a set of labeled data $\mathcal{D}_{L}^{j}$ and a set of unlabeled data $\mathcal{D}_{U}^{j}$. $f_{\mathcal{D}^{j}}$ is the model trained with dataset $\mathcal{D}^{j}$. We consider two cases; 1) $f_{\mathcal{D}^{j}}$ is trained  with only $\mathcal{D}_L^{j}$ via supervised learning and 2) $f_{\mathcal{D}^{j}}$ is trained  with $\mathcal{D}_L^{j} \cup \mathcal{D}_U^{j}$ via SSL. The AL algorithm aims to acquire the set of $b$ data samples $\{x_1^{*},...,x_b^{*}\}$ from the pool of unlabeled data $\mathcal{D}_{U}^{j}$. To select the data samples to be labeled, the acquisition function $a_{\rm ST-CoNAL}\left(x_i, f_{\mathcal{D}^{j}}\right)$ is used to score the given data instance $x_i$ with the currently trained model $f_{\mathcal{D}^{j}}$.  The $b$ samples that yield the highest score are selected as
\begin{align}
            \{x_{1}^{*}, ...,x_{b}^{*}\} = \underset{\{x_{1}, ...,x_{b}\}\subset \mathcal{D}^{j}_{U}} {\operatorname*{argmax}} \sum_{i=1}^{b} a_{\rm ST-CoNAL}\left(x_i, f_{\mathcal{D}^{j}} \right). \label{eq:argmax_acquisition}
\end{align}
The selected samples are then labeled, i.e., $\ \mathcal{D}^{j+1}_{L}\leftarrow \mathcal{D}^{j}_{L}\cup \{x_1^{*},...,x_b^{*}\}$ and $\mathcal{D}^{j+1}_{U} \leftarrow \mathcal{D}^{j}_{U} \setminus \{x_1^{*},...,x_b^{*}\}$, and the model is retrained using the newly labeled dataset $\mathcal{D}^{j+1}=\{\mathcal{D}^{j+1}_{L}, \mathcal{D}^{j+1}_{U}\}$. %
This procedure is repeated until a labeled dataset is obtained as large as required within the limited budget. 

Suppose that we have $Q$ student models $f_{\mathcal{D}^{j}}^{(s,1)},...,f_{\mathcal{D}^{j}}^{(s,Q)}$ and one teacher model $f_{\mathcal{D}^{j}}^{(t)}$ trained with the dataset $\mathcal{D}^{j}$. Each model produces the $K$ dimensional output through the softmax output layer. % $f^{(\cdot)}_{\mathcal{D}^{j}} = \left[f^{(\cdot)}_{\mathcal{D}^{j}, 1}, ..., f^{(\cdot)}_{\mathcal{D}^{j}, K}\right]^T$. 
The method for obtaining these models will be discussed in the next section.
%in Section \ref{consistency measure based on student-teacher}. 
We first apply sharpening \cite{berthelot2019mixmatch, berthelot2019remixmatch, xie2019UDA} to the output of the teacher model as
\begin{align}
       \bar{f}^{(t)}_{\mathcal{D}^{j}}(k) &= \frac{ \left(f^{(t)}_{\mathcal{D}^{j}}(k)\right)^{\frac{1}{T}}}{\sum_{k=1}^{K} \left(f^{(t)}_{\mathcal{D}^{j}}(k)\right)^{\frac{1}{T}}}, 
       \label{eq:sharpening} 
\end{align}
where $f(k)$ denotes the $k$th element of the output produced by the function $f(\cdot)$ and $T$ denotes the temperature hyper-parameter. Then, the acquisition function of the ST-CoNAL is obtained by accumulating the KL divergence between the predictions produced by each student model and the teacher model, i.e.,
\begin{align}
       a_{\text{ST-CoNAL}}(x_i, {f}_{\mathcal{D}^{j}}) &= \frac{1}{Q}\sum_{q=1}^{Q} KL (\bar{f}_{\mathcal{D}^{j}}^{(t)}(x_i)|| f_{\mathcal{D}^{j}}^{(s,q)}(x_i)),
       \label{eq:ST-CoNAL acquisition} 
\end{align}
 where $KL(p||q) = \sum_i p_i \log(p_i/q_i)$. The KL divergence in this acquisition function measures the inconsistency between the student and teacher models.  As the sharpening makes the prediction of teacher model to have low-entropy, the high KL divergence value reflects the uncertainty in the prediction of the student models.

\begin{algorithm}[t]
\caption{ST-CoNAL}
\SetAlgoLined
\DontPrintSemicolon
\SetKwInput{kwinput}{Input}
\kwinput{\\
Initial Dataset: $\mathcal{D}^{1} = (\mathcal{D}^{1}_{L}, \mathcal{D}^{1}_{U})$ \\ 
% Number of classes: $K$ \\
Total number of acquisition steps: $J$ \\
Labeling budget: $b$ \\
Temperature parameter: $T$
}

\For{$j=1$ \KwTo $ J $}{
    {\bf Training:} \\ 
    Generate the student models $f_{\mathcal{D}^{j}}^{(s,1)}, ..., {f_{\mathcal{D}^{j}}^{(s,Q)}}$ using SGD with the modified LR schedule in (4). \\ 
    {\bf Sample acquisition:} \\
    Compute the teacher model $f_{\mathcal{D}^{j}}^{(t)}$ via weight averaging. \\
    $ \bar{f}_{\mathcal{D}^{j}}^{(t)} = \text{Sharpen}(f_{\mathcal{D}^{j}}^{(t)}, T)$. \\ 
    Evaluate $a_{\text{ST-CoNAL}}(x_i, {f}_{\mathcal{D}^{j}})$ for all $x_i \in \mathcal{D}_U^j$. \\
    Acquire $b$ samples according to (\ref{eq:argmax_acquisition}).\\
    {\bf Dataset update:}\\
    $\{(x^*_1, y^*_1), ..., (x^*_b, y^*_b)\} = {\rm Annotate}(\{x^*_1, ..., x^*_b\})$ \\
    $\mathcal{D}^{j+1}_{L} \leftarrow \mathcal{D}^{j}_{L} \cup \{x^*_1, ..., x^*_b\}$\\
    $\mathcal{D}^{j+1}_{U} \leftarrow \mathcal{D}^{j}_{U} \setminus \{x^*_1, ..., x^*_b\}$\\
    $\mathcal{D}^{j+1} \leftarrow (\mathcal{D}^{j+1}_{L}, \mathcal{D}^{j+1}_{U})$\\
}
Train the model using $\mathcal{D}^{J+1}$ for SSL or $\mathcal{D}^{J+1}_L$ for supervised learning. \\
\textbf{return} Trained model\\
\label{algorithm:ST-CoNAL}
\end{algorithm}

\subsection{Generation of Student and Teacher Models via Optimization Path} 
\label{consistency measure based on student-teacher}

The ST-CoNAL computes the student and teacher models using the temporal self-ensemble networks. The parameters of these self-ensemble networks are obtained by capturing the network weights at the intermediate check points of the weight trajectory. The study in \cite{athiwaratkun2018fastswa} revealed that the average of the weights obtained through the SGD iterates can yield the weights corresponding to the flat minima of the loss surface, which are known to generalize well \cite{maddox2019simple, guandao2018swalp}.  In ST-CoNAL, the average of these weights forms a teacher model. To obtain the diverse student models, we adopt the learning rate (LR) schedule used for SWA \cite{izmailov2018averagingswa, maddox2019simple, guandao2018swalp}.
The learning rate $l$ used for our method is given by 
\begin{equation}
    l = \begin{cases}
    l_0, & \text{if $t< T_0$}\\
    \gamma l_0, & \text{otherwise}
    \end{cases}
\end{equation}
where $l_0$ denotes the initial learning rate, $\gamma < 1$ denotes the parameter for learning rate decay, $t$ is the current training epoch, and $T_0$ denotes the training epoch at which the learning rate will be switched to a smaller value.
After running $T_0$ epochs (i.e., $t \geq T_0$), the $Q$ intermediate network weights are stored every $c$ epoch, which constitute the weights of the student networks. We obtain $Q$ self-ensemble networks $f_{\mathcal{D}^{j}}^{(s, 1)},...,f_{\mathcal{D}^{j}}^{(s, Q)}$  and these models are treated as the student models. The teacher model $f_{\mathcal{D}^{j}}^{(t)}$ is then obtained by taking the equally weighted average of the parameters of  $f_{\mathcal{D}^{j}}^{(s, 1)},...,f_{\mathcal{D}^{j}}^{(s, Q)}$. Note that this learning rate schedule does not require any additional training cost for obtaining the temporal self-ensemble.

\subsection{Summary of Proposed ST-CoNAL}
\label{summary of ST-CONAL}
Algorithm \ref{algorithm:ST-CoNAL} presents the summary of the ST-CoNAL algorithm.

\section{Experiments}
\label{sec:experiments}
In this section, we evaluate the performance of ST-CoNAL via experiments  conducted on four public datasets, CIFAR-10 \cite{krizhevsky2009learning}, CIFAR-100 \cite{krizhevsky2009learning}, Caltech-256 \cite{griffin2007caltech} and Tiny ImageNet \cite{le2015tiny}. 

\subsection{Experiment Setup}
{\bf Datasets.}
We evaluated ST-CoNAL on four benchmarks: CIFAR-10 \cite{krizhevsky2009learning}, CIFAR-100 \cite{krizhevsky2009learning}, Caltech-256 \cite{griffin2007caltech} and Tiny ImageNet \cite{le2015tiny}.  CIFAR-10 contains $50k$ training examples with 10 categories and CIFAR-100 contains $50k$ training examples with 100 categories.
Caltech-256 has $24,660$ training examples with 256 categories and Tiny ImageNet has $100k$ training examples with 200 categories.

To see how ST-CoNAL performed on class-imbalanced scenarios, we additionally used the synthetically imbalanced CIFAR-10 datasets; {\it the step imbalanced CIFAR-10} \cite{kim2021task, cui2019class} and the {\it long-tailed CIFAR-10} \cite{cao2019learning, cui2019class}. The step imbalance CIFAR-10 has 50 samples for the first five smallest classes and 5,000 samples for the last five largest classes. 
Long-tailed CIFAR-10 followed the configuration used in \cite{cao2019learning}, where 12,406 training images were generated with 5,000 samples from the largest class and 50 samples from the smallest class. The imbalance ratio of 100 is the widely used setting adopted in the literature \cite{kim2021task, cao2019learning}.

\noindent {\bf Implementation Details.}
In all experiments, an 18-layer residual network (ResNet-18) was used as the backbone network. The detailed structure of this network was provided in \cite{he2016deep}. For CIFAR-10 and CIFAR-100 datasets, models were trained with standard data augmentation operations including random translation and random horizontal flipping \cite{he2016deep}. For Caltech-256 and Tiny-ImageNet datasets, we applied the data augmentation operations including resizing to 256$\times$256, random cropping to 224$\times$224, and horizontal flipping. In the inference phase, we applied the image resizing method followed by the center cropping to generate 224$\times$224 input images \cite{he2016deep}. We also applied our ST-CoNAL method to SSL setup. We used the mean teacher method using the detailed configurations described in \cite{tarvainen2017meanteacher}. We conducted all experiments with a TITAN XP GPU for training.

\noindent {\bf Configurations  for AL.}
According to  \cite{sener2017coreset, settles2009activesurvey}, selection of $b$ samples from a large pool of unlabeled samples requires computation of the acquisition function for all unlabeled samples in the pool, yielding considerable computational overhead. To mitigate this issue, the subset of unlabeled samples $\mathcal{S}$ was randomly drawn from the unlabeled pool $\mathcal{D}_{U}$   \cite{beluch2018thepower, yoo2019learning, kim2021task}. Then, acquisition functions were used to acquire the best $b$ samples from $\mathcal{S}$. 

In all experiments, the temperature parameter $T$ was set to $0.7$, which was determined through empirical optimization. 
For CIFAR-10, we set $\gamma=1.0, c=10, b=1k$  and $|\mathcal{S}|= 10k$,  and for CIFAR-100, we set $\gamma=0.5, c=10, b=2k$  and $|\mathcal{S}|= 10k$. For Tiny ImageNet, we set $\gamma=0.5, c=10, b=2k$ and $|\mathcal{S}|= 20k$. For Caltech-256, we set $\gamma=0.3, c=10, b=1k$ and $|\mathcal{S}|=10k$. These configurations followed the convention adopted in the existing methods.  The parameter $T_0$ used for the learning rate scheduling was set to 160 epochs for all datasets except Tiny ImageNet. We used $T_0$ of 60 epochs for  Tiny ImageNet. 

%In the initial acquisition step where the trained model is not available, the labeled data was randomly selected from the dataset.

\begin{figure*}[t]
    % \centering
    \begin{subfigure}[b]{0.45\textwidth}
        \centering
        \includegraphics[width=1.0\textwidth]{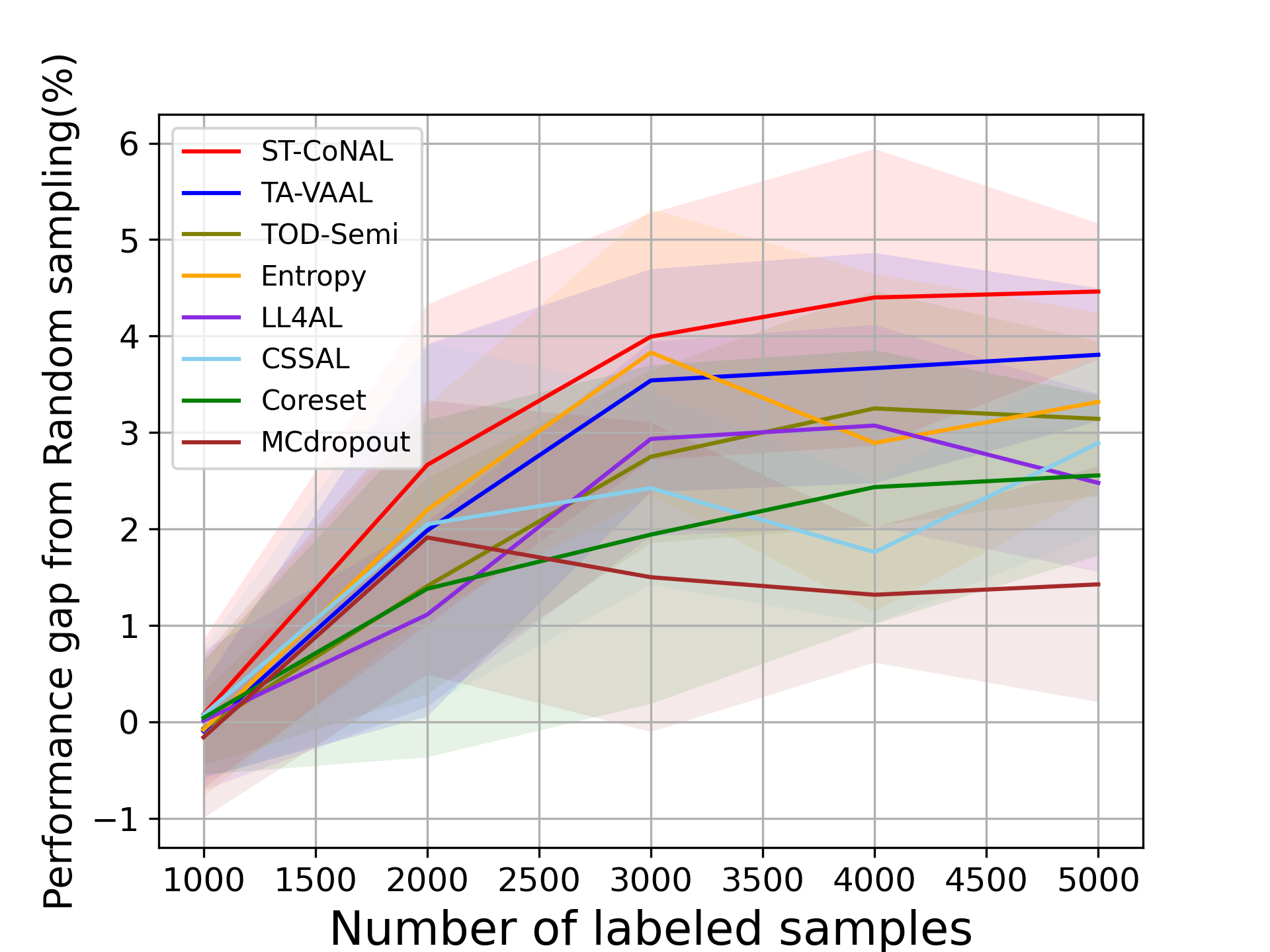}
        \caption[]{}
        \label{fig:cifar-10 sl}
    \end{subfigure}
    \hfill
    \begin{subfigure}[b]{0.45\textwidth}  
        \centering 
        \includegraphics[width=1.0\textwidth]{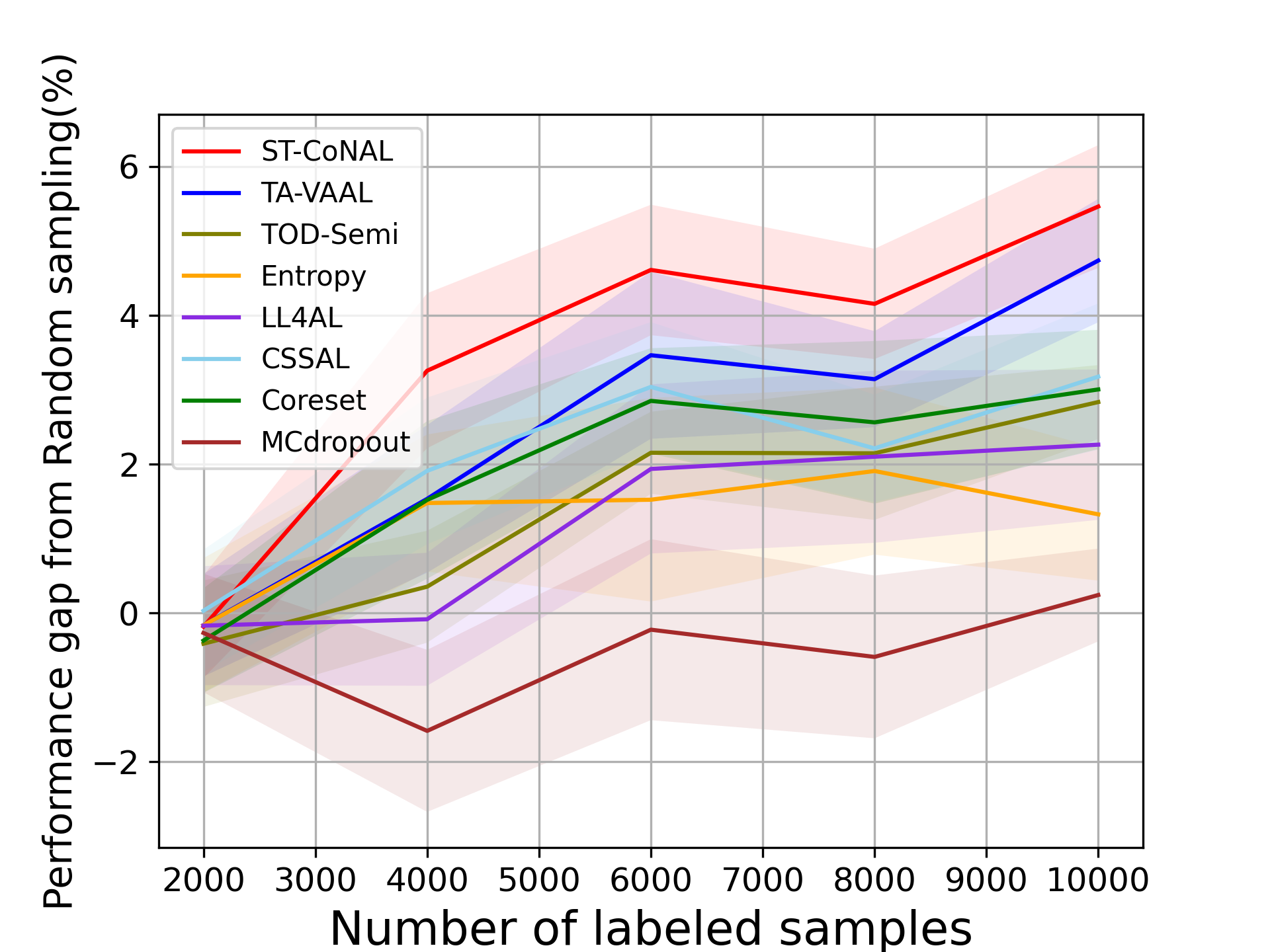}
        \caption[]{}
        \label{fig:cifar-100 sl}
    \end{subfigure}
    \captionsetup{singlelinecheck = false}
    \caption[]
    {Average accuracy improvement from random sampling versus the number of labeled samples on (a) CIFAR-10 and (b) CIFAR-100 datasets.}
    \label{fig:cifar sl}
\end{figure*}

\begin{figure*}[t]
    % \centering
    \begin{subfigure}[b]{0.45\textwidth}
        \centering
        \includegraphics[width=1.0\textwidth]{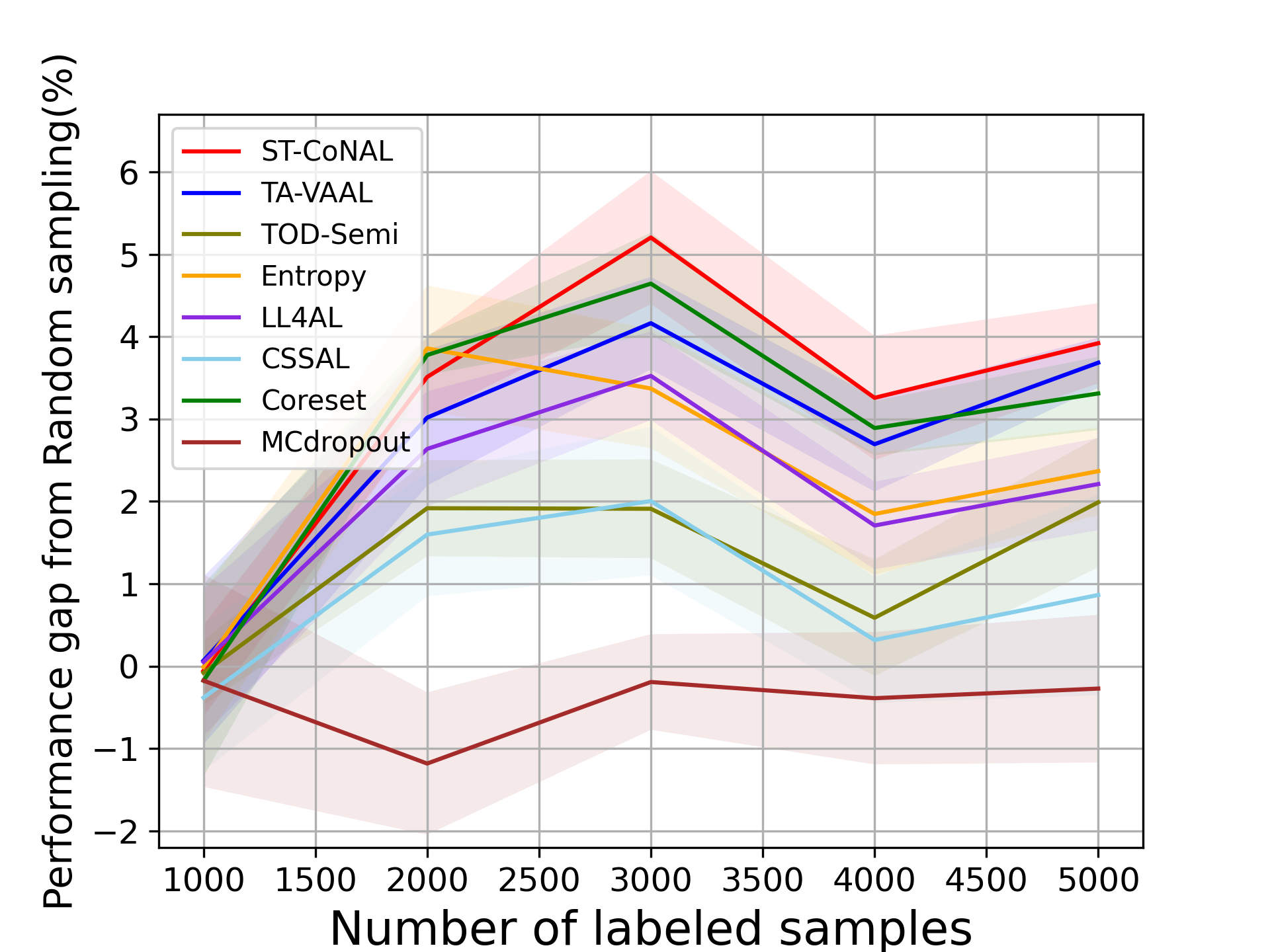}
        \caption[]{}
        \label{fig:caltech sl ema}
    \end{subfigure}
    \hfill
    \begin{subfigure}[b]{0.45\textwidth}  
        \centering 
        \includegraphics[width=1.0\textwidth]{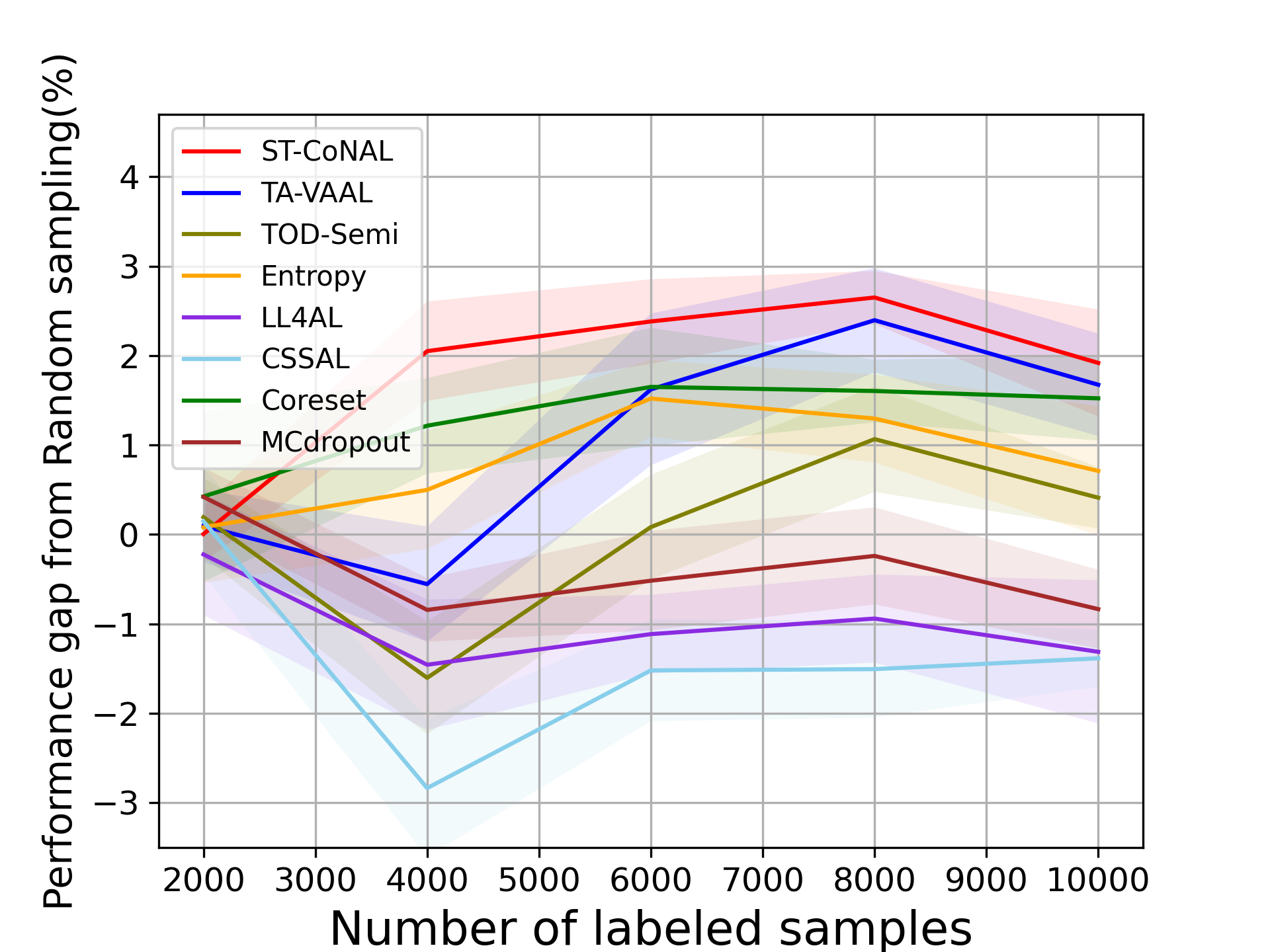}
        \caption[]{}
        \label{fig:tiny sl ema}
    \end{subfigure}
    \captionsetup{singlelinecheck = false}
    \caption[]
    {Average accuracy improvement from random sampling versus the number of labeled samples on (a) Caltech-256 and (b) Tiny ImageNet datasets.}
    \label{fig:caltech tiny sl}
\end{figure*}

\begin{figure*}[tb]
    % \centering
    \begin{subfigure}[b]{0.45\textwidth}
        \centering
        \includegraphics[width=1.0\textwidth]{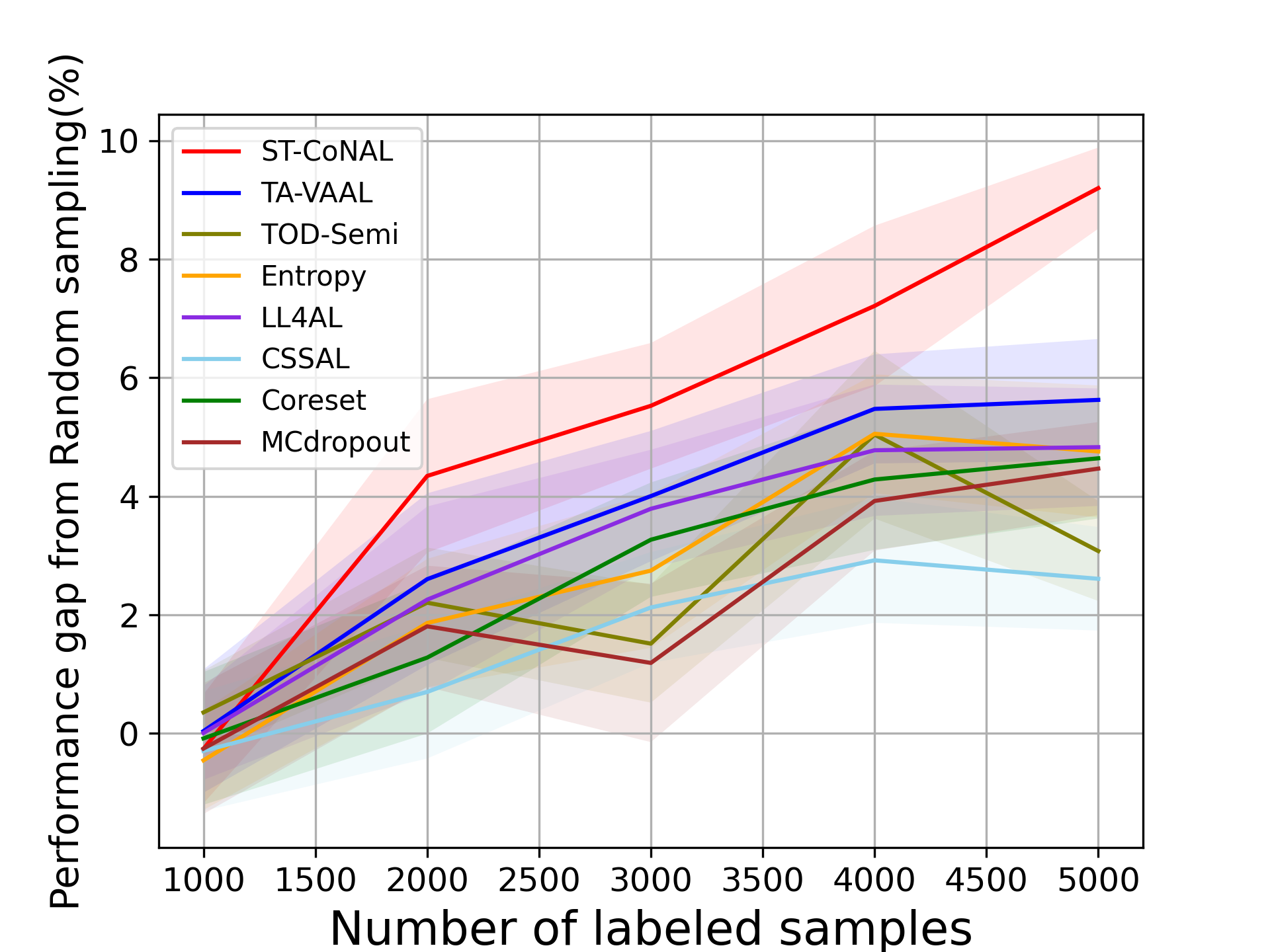}
        \caption[]{}
        \label{fig:cifar-10 imbalance}
    \end{subfigure}
    \hfill
    \begin{subfigure}[b]{0.45\textwidth}  
        \centering 
        \includegraphics[width=1.0\textwidth]{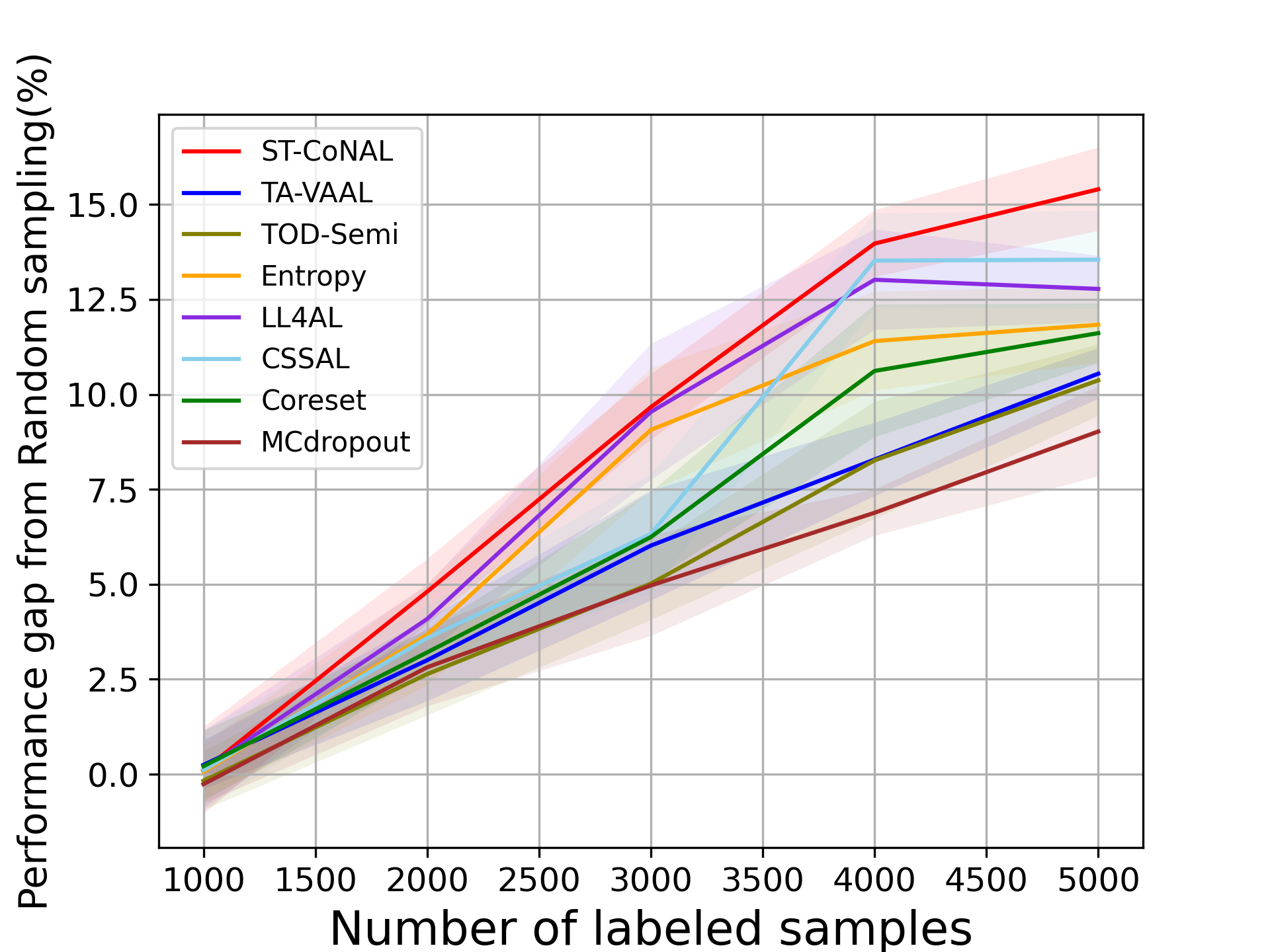}
        \caption[]{}
        \label{fig:cifar-100 imbalance}
    \end{subfigure}
    \captionsetup{singlelinecheck = false}
    \caption[]
    {Average accuracy improvement from random sampling  versus the number of labeled samples on (a) the step imbalanced CIFAR-10 and (b) the long-tailed CIFAR-10. The imbalance ratio was set to 100 for both experiments.}
    \label{fig:cifar imbalance}
\end{figure*}

\begin{figure*}[tb]
    % \centering
    \begin{subfigure}[b]{0.45\textwidth}
        \centering
        \includegraphics[width=1.0\textwidth]{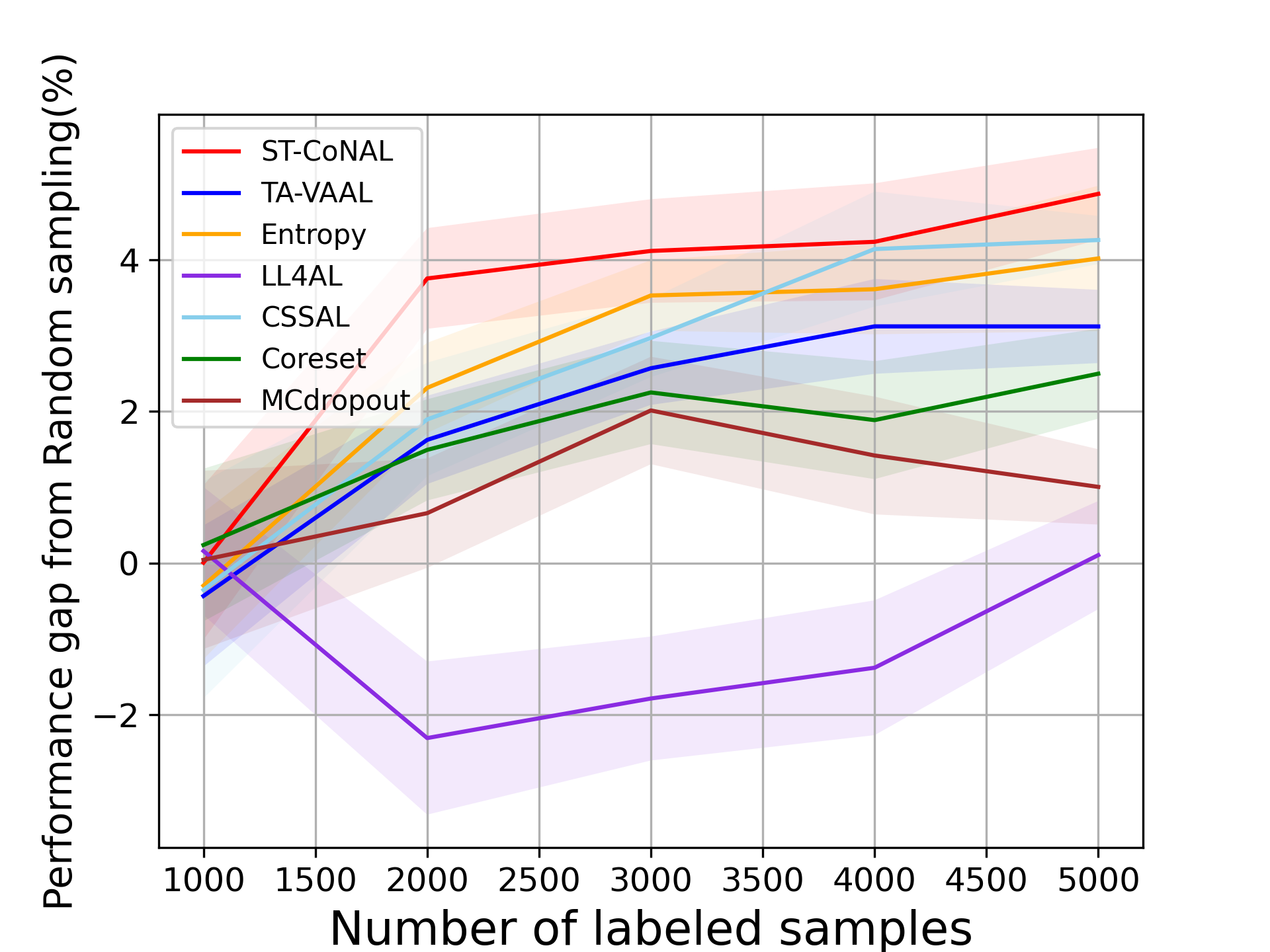}
        \caption[]{}
        \label{fig:cifar-10 ssl ema}
    \end{subfigure}
    \hfill
    \begin{subfigure}[b]{0.45\textwidth}  
        \centering 
        \includegraphics[width=1.0\textwidth]{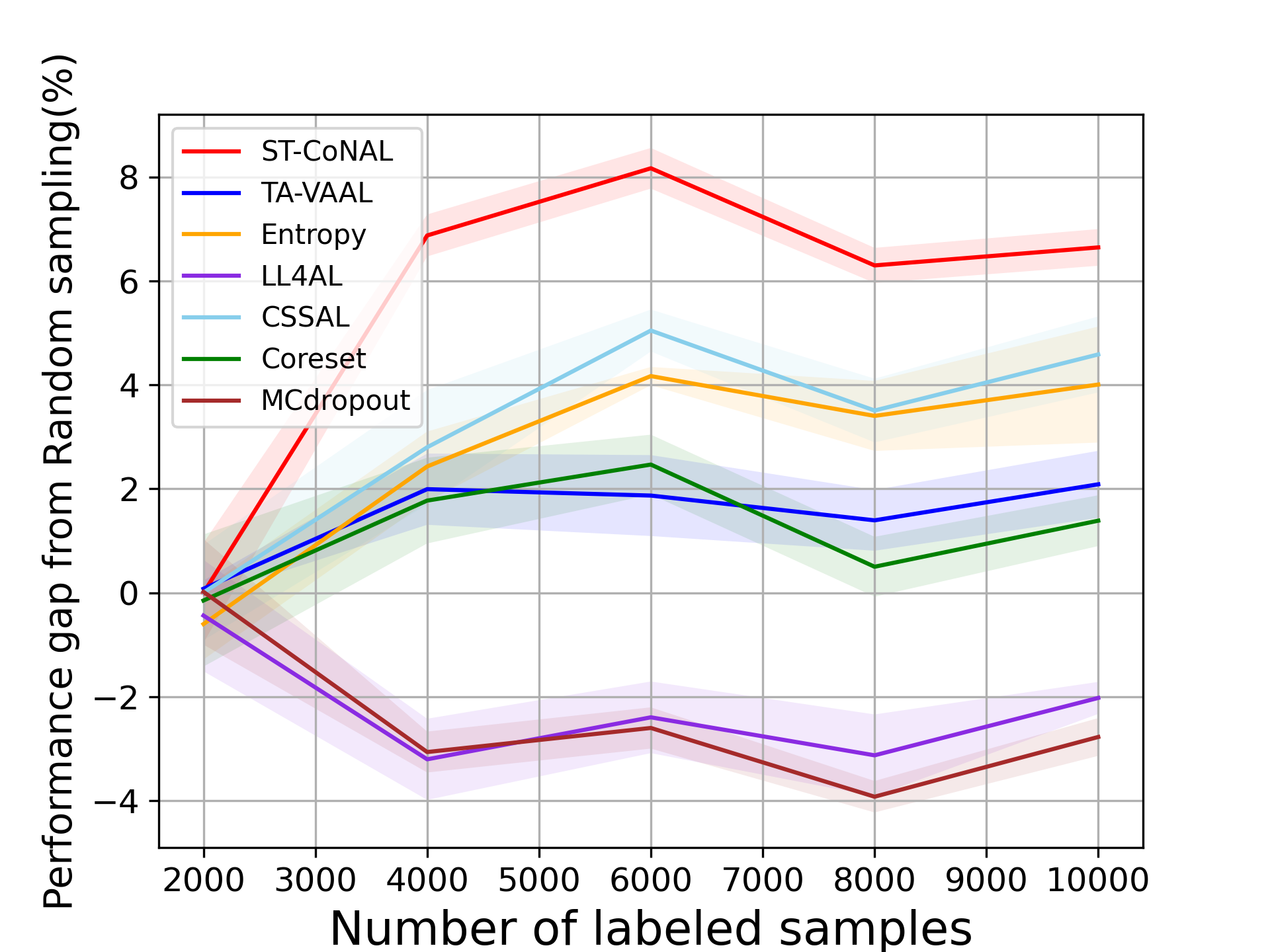}
        \caption[]{}
        \label{fig:cifar-100 ssl ema}
    \end{subfigure}
    \captionsetup{singlelinecheck = false}
    \caption[]
    {Average accuracy improvement from random sampling versus the number of labeled samples on (a) CIFAR-10 and (b) CIFAR-100 dataset. The model was trained by the mean teacher \cite{tarvainen2017meanteacher}, one of SSL methods.}
    \label{fig:cifar ssl}
\end{figure*}

\noindent {\bf Baselines} We compared our method with the existing AL methods including Core-set \cite{sener2017coreset}, MC-Dropout \cite{gal2016dropout}, LL4AL \cite{yoo2019learning}, Entropy \cite{shannon1948mathematical}, CSSAL \cite{gao2020consistency} , TOD-Semi \cite{huang2022temporalTOD}, TA-VAAL \cite{kim2021task}, and random sampling. We adopted the average classification accuracy as a performance metric. We report the mean and standard deviation of the classification accuracy measured 5 times with different random seeds. In our performance comparison, we kept the network optimization and inference methods the same and only changed the acquisition criterion. TOD-Semi, TA-VAAL and LL4AL are exceptions because we used the optimization and inference methods proposed in their original paper. In Fig. \ref{fig:cifar sl} to  Fig. \ref{fig:cifar ssl}, the performance difference of the target method from the random sampling baseline was used as a performance indicator.

\subsection{Performance Comparison}
{\bf CIFAR-10 and CIFAR-100: } Fig. \ref{fig:cifar sl} (a) and (b) show the performance of ST-CoNAL with respect to the number of labeled samples evaluated on CIFAR-10 and CIFAR-100 datasets, respectively. As mentioned earlier, we use the performance difference obtained compared to the classification accuracy of random sampling as a performance indicator. We observe that the proposed ST-CoNAL outperforms other AL methods on both CIFAR-10 and CIFAR-100. After the last acquisition step, our method achieves a performance improvement of 4.68\% and 5.47\% compared to the random sampling baseline on CIFAR-10 and CIFAR-100 datasets, respectively. It should be noted that ST-CoNAL achieves performance comparable to TA-VAAL, a state-of-the-art method that requires the addition of subnetworks and complex training optimization. Entropy shows a good performance at the beginning acquisition steps but the performance improves slowly as compared to other methods. After the last acquisition, ST-CoNAL achieves 1.37\% and 4.15\% higher accuracy than Entropy on CIFAR-10 and CIFAR-100, respectively. Our method outperforms CSSAL, another consistency-based AL method that only considers data uncertainty, not model uncertainty.

{\bf Caltech-256 and Tiny ImageNet: } 
Fig. \ref{fig:caltech tiny sl} (a) and (b) show the performance of several AL methods on Caltech-256 and Tiny ImageNet, respectively. The ST-CoNAL also achieves significant performance gain on these datasets.  
After the last acquisition step, ST-CoNAL achieves 3.92\% higher accuracy than the random sampling baseline on Caltech-256 while it achieves 1.92\% gain on Tiny-Imagenet dataset. Note that the proposed method far outperforms the CSSAL and performs comparable to more sophisticated representation-based methods, Coreset and TA-VAAL.

\subsection{Performance Comparison on Other Setups }

{\bf Class-Imbalanced Datasets.} We also evaluate ST-CoNAL on class imbalanced datasets. Fig. \ref{fig:cifar imbalance} (a) and (b) present the performance of ST-CoNAL on (a) the step imbalanced  CIFAR-10 \cite{kim2021task} and (b) the long-tailed CIFAR-10  \cite{cao2019learning}.  We see that ST-CoNAL achieves remarkable performance improvements compared to random sampling on the class imbalanced datasets. Specifically, after the last acquisition step, ST-CoNAL achieves a 9.2\% performance improvement on the step-imbalanced CIFAR-10 and a 15.41\% improvement on the long-tailed CIFAR-10. 
Furthermore, ST-CoNAL achieves larger performance gain over the current state-of-the-art, TA-VAAL. It achieves  3.58\% and 4.86\% performance gains over TA-VAAL on the step imbalanced CIFAR-10 and the long-tailed CIFAR-10, respectively. Note that the  consistency-based method CSSAL does not perform well under this class-imbalanced setup.

{\bf SSL Setups.} Fig. \ref{fig:cifar ssl} (a) and (b) show the performance evaluated on CIFAR-10 and CIFAR-100 datasets when ST-CoNAL is applied to SSL. ST-CoNAL maintains performance gains even in SSL settings and outperforms other AL methods by a larger margin than in supervised learning settings. After the last acquisition step, ST-CoNAL achieves the 4.87\% and 6.66\% performance gains over the random sampling baseline on CIFAR-10 and CIFAR-100, respectively. In particular, Fig. \ref{fig:cifar ssl} (b) shows that ST-CoNAL achieves substantial performance gains over the existing methods in CIFAR-100.

\begin{figure*}[t]
    \begin{subfigure}[]{0.45\textwidth}   
        \centering 
        \includegraphics[width=1\textwidth,height=0.475\textwidth]{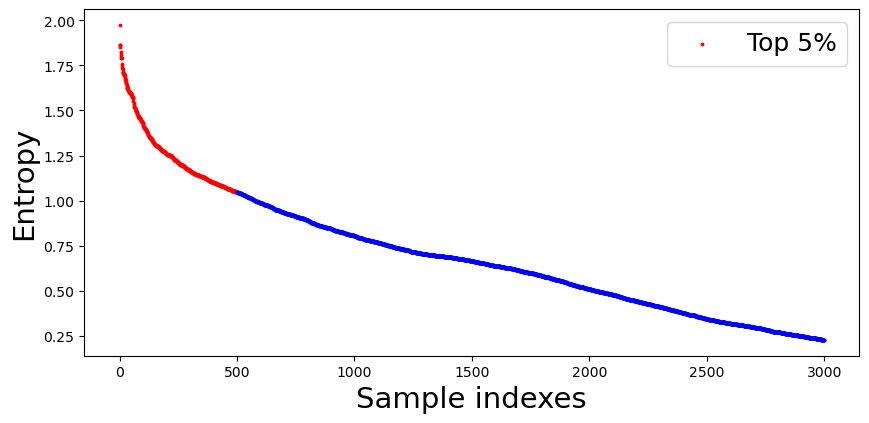}
        \caption[]{}
        \label{fig:cifar10_conf_by_conf}
    \end{subfigure}
    \hfill
    \begin{subfigure}[]{0.45\textwidth}   
        \centering 
        \includegraphics[width=1\textwidth,height=0.475\textwidth]{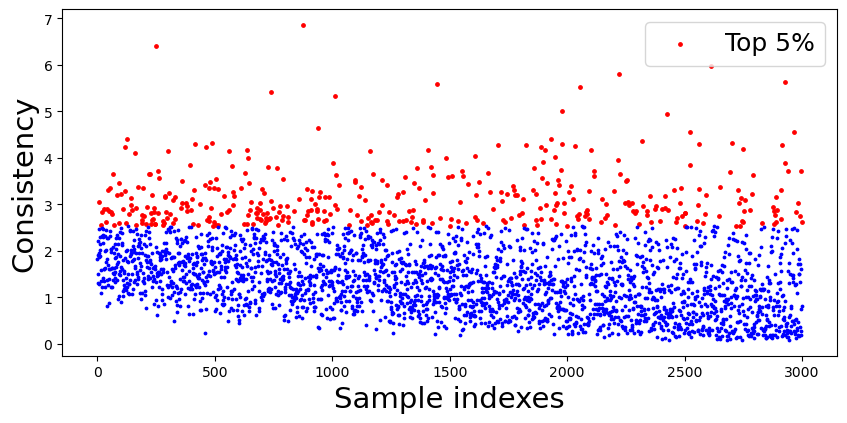}
        % \includesvg[width=1\textwidth]{figure/exp_ST-CoNAL_based_try_2_8/consistency_measure.svg}
        \caption[]{}
        %{{\small  Consistency measure in descending order of confidence measure}} 
        \label{fig:cifar10_cons_by_conf}
    \end{subfigure}
    \captionsetup{singlelinecheck = false}
    \caption[]
    {The entropy and consistency measure are evaluated based on $10k$ unlabeled samples. The model is trained after $4k$ samples are acquired on CIFAR-10 dataset. (a) Entropy evaluated for the samples sorted in descending order by entropy and (b) consistency measure evaluated for the samples sorted in the same order. Top 5\% of samples are marked in red. }
    \label{fig:cifar10_acquisition_value_by_criterion}
\end{figure*}

\begin{figure}[t]
    \centering
    \includegraphics[width=0.6\textwidth, height=0.3\textwidth]{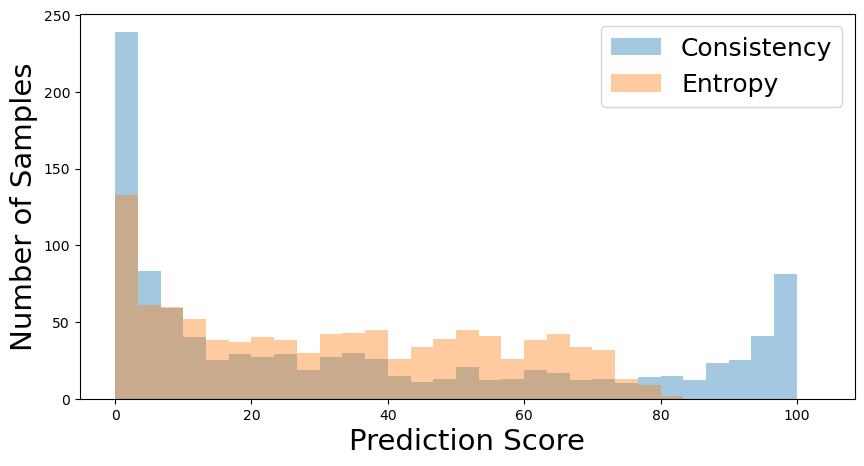}
    \caption[]{Histogram of prediction scores produced by the samples selected by the proposed acquisition criterion and entropy-based one. }
    \label{fig:prediction probability}
\end{figure}

\subsection{Performance Analysis}
In this subsection, we investigate the benefit of using self-ensemble for an acquisition criterion through some experiments. 

\label{analysis of consistency measure}
\subsubsection{Comparison Between Consistency Measure versus Uncertainty Measure.} \label{cop}
To understand the behavior of the proposed consistency measure, we compare it with the popularly used uncertainty measure, Entropy \cite{shannon1948mathematical}. The sample acquisition process was conducted with the setup described in {\it Experiment Setup} section. We stopped the acquisition when $4k$ samples are labeled. Then, we evaluated both the entropy measure and the consistency measure of ST-CoNAL with 10k samples randomly selected from the remaining unlabeled samples.  The 10k samples were sorted in descending order according to entropy and sample indexes were assigned in order. Fig. \ref{fig:cifar10_acquisition_value_by_criterion} (a) shows the plot of entropy value versus the sample index. Obviously, due to sorting operation, the entropy value decreases with the sample index.  Fig. \ref{fig:cifar10_acquisition_value_by_criterion} (b) presents the plot of the consistency measure over the samples sorted in the same order. The samples ranked in the top 5\% by each measure are marked in red in each figure. We observe that quite different samples were identified in the top 5\% according to two criteria. From this, we can conclude that the consistency measure offers the quite different standard for acquisition from the entropy measure.

 Fig. \ref{fig:prediction probability} shows the distribution of the prediction scores produced by the samples selected by the proposed acquisition criterion and the conventional entropy-based one \cite{shannon1948mathematical}. We use a prediction score that corresponds to the true class index for the given sample. We used the same setup in Fig. \ref{fig:cifar10_acquisition_value_by_criterion}. Obviously, the entropy-based criterion tends to select the samples with low confidence. It does not select the samples whose prediction score is above 80\%. On the contrary, the proposed acquisition criterion selects samples with widely spread prediction scores. Note that it even selects samples for which predictions were made with near 100\% confidence because they yielded the low consistency measure.  The fact that the proposed method outperforms entropy-based acquisitions indicates that selecting samples with the lowest confidence is not necessarily the best strategy, and consistency-based acquisitions could promise a better solution by leveraging the advantage of various temporal self-ensemble models.

\begin{table*}[t]
  \centering
  \begin{adjustbox}{width=0.95\columnwidth, center}
  \begin{tabular}{c|ccccc}
  \hline\hline
  CIFAR-10 & $1k$ samples labeled & $2k$ samples labeled & $3k$ samples labeled & $4k$ samples labeled & $5k$ samples labeled\\
  \hline
  EMA & 46.36 & 60.73 & 69.10 & 77.49 & 81.51\\
  Our EWA &   47.13 & \bf{61.48} & \bf{71.85} & \bf{79.61} & \bf{83.05} \\
  \hline\hline
  CIFAR-100 & $2k$ samples labeled & $4k$ samples labeled & $6k$ samples labeled & $8k$ samples labeled & $10k$ samples labeled\\
  \hline
  EMA & 20.34 & 32.96 & 42.75  &51.77 & 55.49 \\
  Our EWA & 20.35 & \bf{34.70} & \bf{44.87} & \bf{52.47} & \bf{57.49} \\
  \hline
  \hline
  \end{tabular}
  \end{adjustbox}
  \caption{Ablation study on the equally-weighted average (EWA) strategy compared to exponential moving average (EMA) of MT \cite{tarvainen2017meanteacher}.}
  \label{table:ewa vs ema}
\end{table*}

\subsubsection{Comparison Between Equally-Weighted  Average versus Exponential Moving Average.}
Table \ref{table:ewa vs ema} compares two acquisition functions that construct a teacher model using the equally-weighted average (EWA)  of the proposed method versus the exponential moving average (EMA) of the MT method. For fair comparison, only the teacher models were differently constructed  while  all other configurations were set equally for both training and inference. 
Table \ref{table:ewa vs ema} shows that EWA achieves higher performance gains over EWA as the acquisition step proceeds on both CIFAR-10 and CIFAR-100 datasets. This shows that the proposed method can construct more effective teacher model for sample acquisition than EMA.

\begin{figure*}[t]
    \begin{subfigure}[]{0.45\textwidth}   
        \centering 
        \includegraphics[width=1.0\textwidth,height=0.80\textwidth]{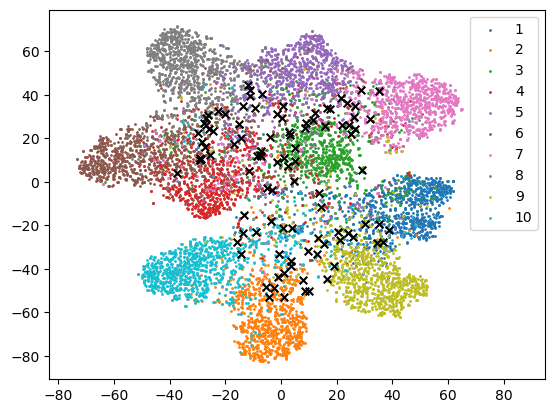}
        \caption[]{}
        \label{fig:tsne entropy}
    \end{subfigure}
    \hfill
    \begin{subfigure}[]{0.45\textwidth}   
        \centering 
        \includegraphics[width=1.0\textwidth,height=0.80\textwidth]{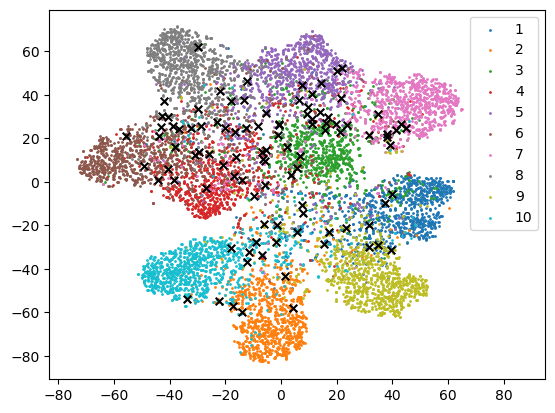}
        \caption[]{}
        \label{fig:tsne consistency}
    \end{subfigure}
    \captionsetup{singlelinecheck = false}
    \caption[]
    {T-SNE \cite{maaten2008visualizing} visualizes the feature at the penultimate layer of the teacher model.  The model is trained by $4k$ samples on CIFAR-10 dataset. The colored dots correspond to the unlabeled samples assigned to each true class and the black cross marks in (a) and (b) correspond to the top 1\% sample selected by the entropy and the consistency measure, respectively.}
    \label{fig:tsne}
\end{figure*}

\subsubsection{T-SNE Visualization. } We also visualize the samples acquired according to the entropy and the proposed consistency measure using T-SNE \cite{maaten2008visualizing}. T-SNE projects the feature vectors onto the embedding space of lower dimension for visualization. The setup used in Fig. \ref{fig:cifar10_acquisition_value_by_criterion} was also used to perform T-SNE.  In Fig. \ref{fig:tsne}, the colored dots correspond to the unlabeled sample points projected in the embedding space. The samples selected by the entropy and by the consistency measure are marked by black cross in Fig. \ref{fig:tsne} (a) and (b), respectively.  We observe from Fig. \ref{fig:tsne} that the samples selected by two acquisition criteria have different distribution, which is consistent with our finding in previous section. Samples selected by entropy tend to gather near the boundary, while samples selected by consistency measure tend to spread further. This shows that more diverse samples are selected according to consistency measure than entropy.

\section{Conclusions}
In this study, we proposed a new acquisition function for pool-based AL, which used the temporal self-ensemble generated during the SGD optimization. The proposed sample acquisition criterion measures the discrepancy between the predictions from the student model and the teacher model, where the $Q$ student models are provided as the self-ensemble models and the teacher model is obtained through a weighted average of the student models. Using the consistency measure based on KL divergence, the proposed ST-CoNAL can acquire better samples for AL. We also showed that the sharpening operation applied to the logits of the teacher model can further improve the AL performance. The experiments conducted on CIFAR-10, CIFAR-100, Caltech-256 and Tiny ImageNet datasets demonstrated that ST-CoNAL achieved the significant performance gains over the existing AL methods. Furthermore, our numerical analysis indicated that our ST-CoNAL performs better than other AL baselines in class-imbalanced and semi-supervised learning  scenarios.

\bibliographystyle{splncs04}
\bibliography{egbib}

\clearpage
\appendix

% \title{Supplementary material for ST-CoNAL: Consistency-Based Acquisition Criterion Using Temporal Self-Ensemble for Active Learning}
% \author{Jae Soon Baik \and
% In Young Yoon \and Jun Won Choi\thanks{Corresponding Author}}
% \institute{Hanyang University, Seoul, Korea\\
% \email{\{jsbaik, inyoungyoon\}@spa.hanyang.ac.kr,}  \email{junwchoi@hanyang.ac.kr}}

% \maketitle
\section*{Appendix}
\section{Experimental Setup}
In this section, we provide the detailed experimental setups. The configuration parameters used for ST-CoNAL are provided in Table \ref{tab:training details}. As for SSL setup, we followed the configurations used in \cite{tarvainen2017meanteacher}. 

\begin{table*}[tbh]
    \centering
    \begin{adjustbox}{width=0.95\columnwidth, center}
    \begin{tabular}{l|cccc}
        \hline
        \hline
        \centering
        Dataset &  CIFAR-10 & CIFAR-100 & Caltech-256 & Tiny ImageNet\\
        \hline
        Initial learning rate $(l_0)$& 0.1& 0.1& 0.01 & 0.005 \\
        Nesterov momentum & 0.9 & 0.9 & 0.9 & 0.9 \\
        Weight decay & 0.0004 & 0.0004 & 0.001 & - \\
        Batch size & 128 & 128 & 128 & 128 \\
        Labeled batch size & 32 & 32 & 32 & 32 \\
        Total epochs & 200 & 200 & 200 & 100 \\
        Size of unlabeled subset $(|\mathcal{S}|)$ & 10k & 20k & 10k & 20k \\
        Budget $(b)$ & 1k & 2k & 1k & 2k \\
        Size of initially labeled set & 1k & 2k & 1k & 2k \\
        Storing interval $(c)$ & 10& 10 & 10 & 10\\
        Learning rate decay $(\gamma)$ & 1.0 & 0.5 & 0.3 & 0.5 \\
        Learning rate decay point $(T_{0})$ & 160 & 160 & 160 & 60 \\
        
        \hline
        \hline
    \end{tabular}
    \end{adjustbox}
    \caption{The configuration parameters used for ST-CoNAL.}
    \label{tab:training details}
\end{table*}

\section{Experimental Results}
\begin{figure*}[tbh]
    % \centering
    \begin{subfigure}[b]{0.490\textwidth}
        \centering
        \includegraphics[width=1.0\textwidth]{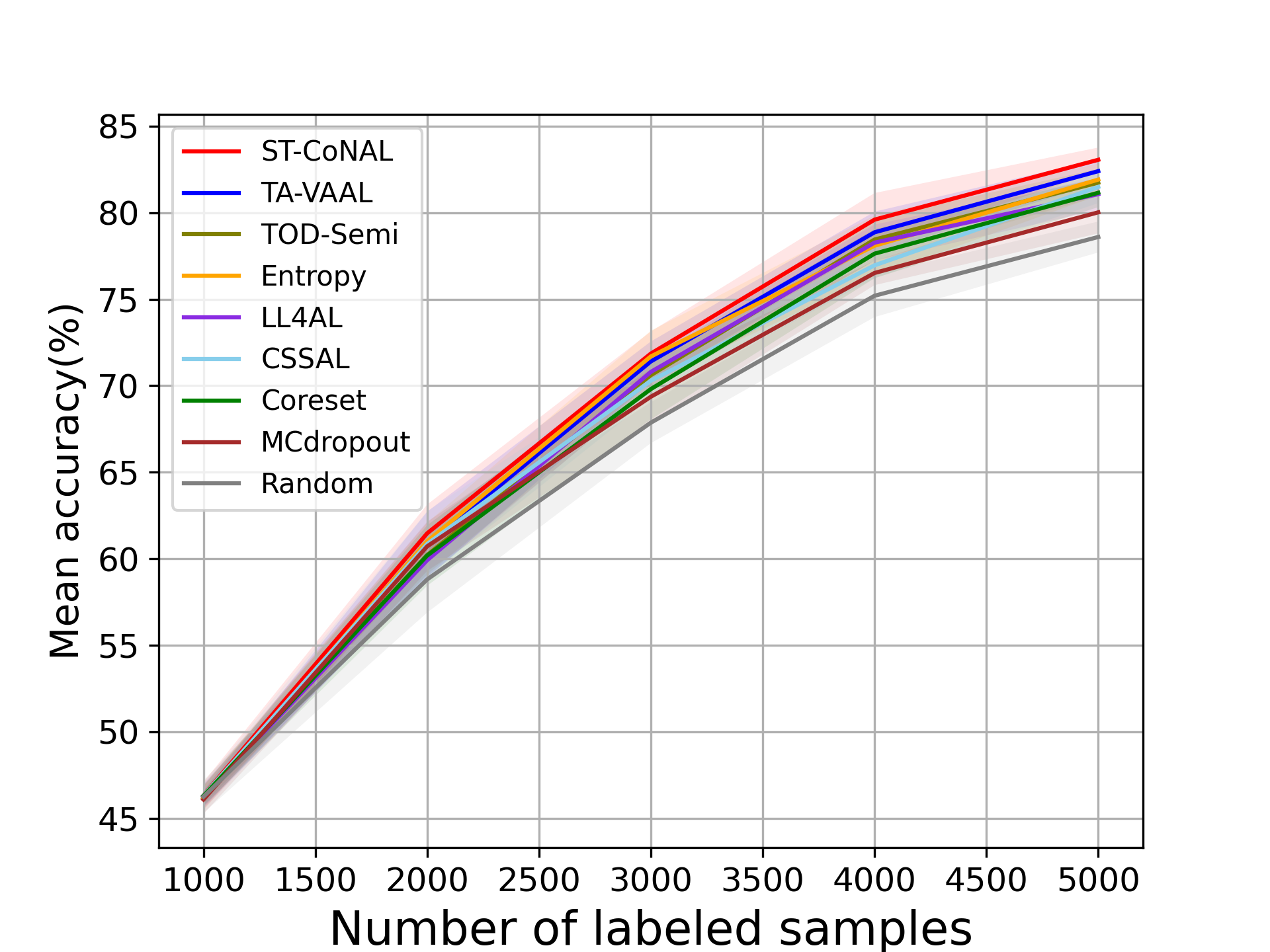}
        \caption[]{}
        \label{fig:cifar-10 sl absolute}
    \end{subfigure}
    \hfill
    \begin{subfigure}[b]{0.490\textwidth}  
        \centering 
        \includegraphics[width=1.0\textwidth]{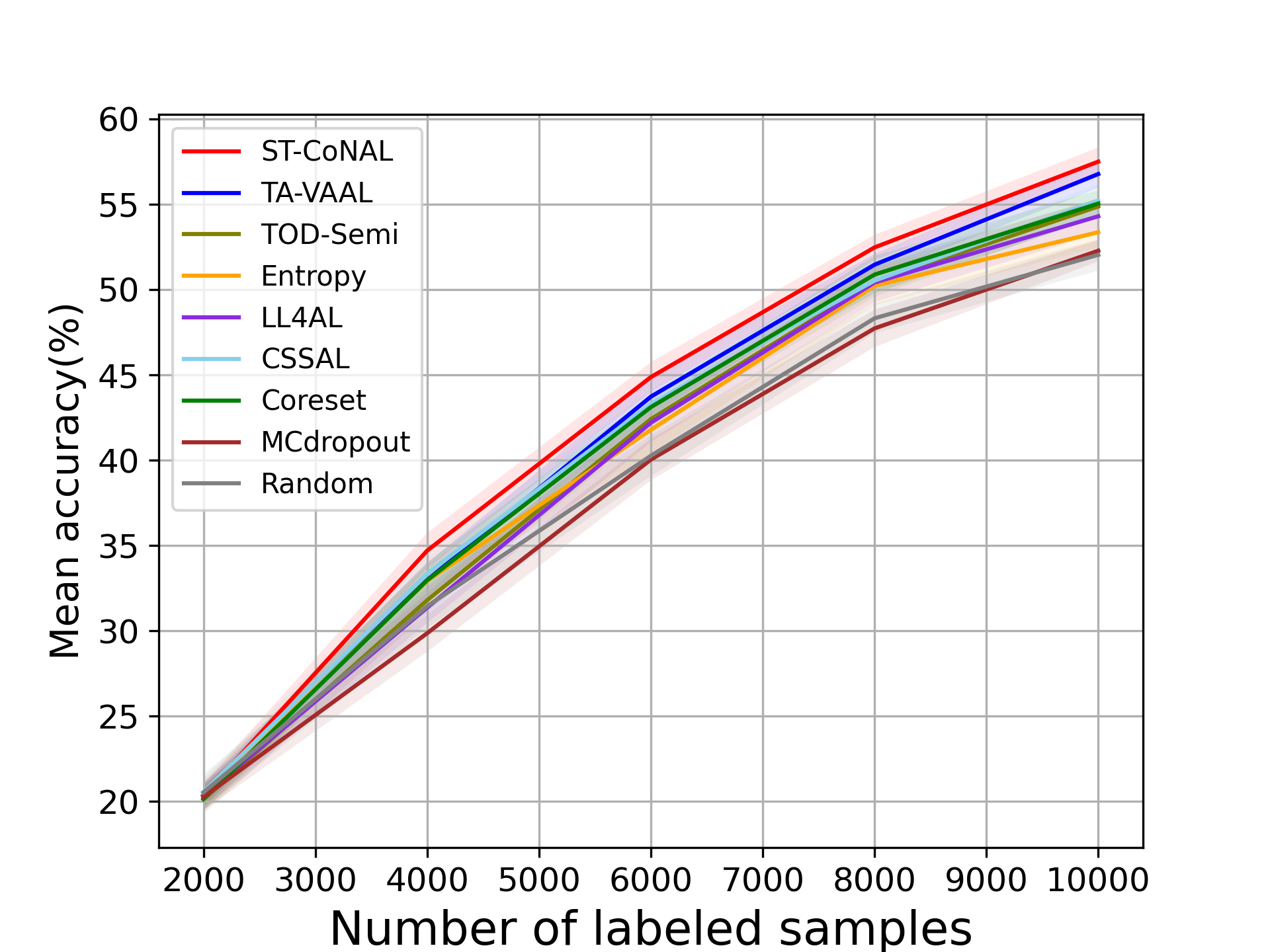}
        \caption[]{}
        \label{fig:cifar-100 sl absolute}
    \end{subfigure}
    \captionsetup{singlelinecheck = false}
    \caption[]
    {Average test accuracy versus the number of labeled samples on (a) CIFAR-10 and (b) CIFAR-100 dataset.}
    \label{fig:cifar sl absolute}
\end{figure*}

\begin{figure*}[tbh]
    % \centering
    \begin{subfigure}[b]{0.490\textwidth}
        \centering
        \includegraphics[width=1.0\textwidth]{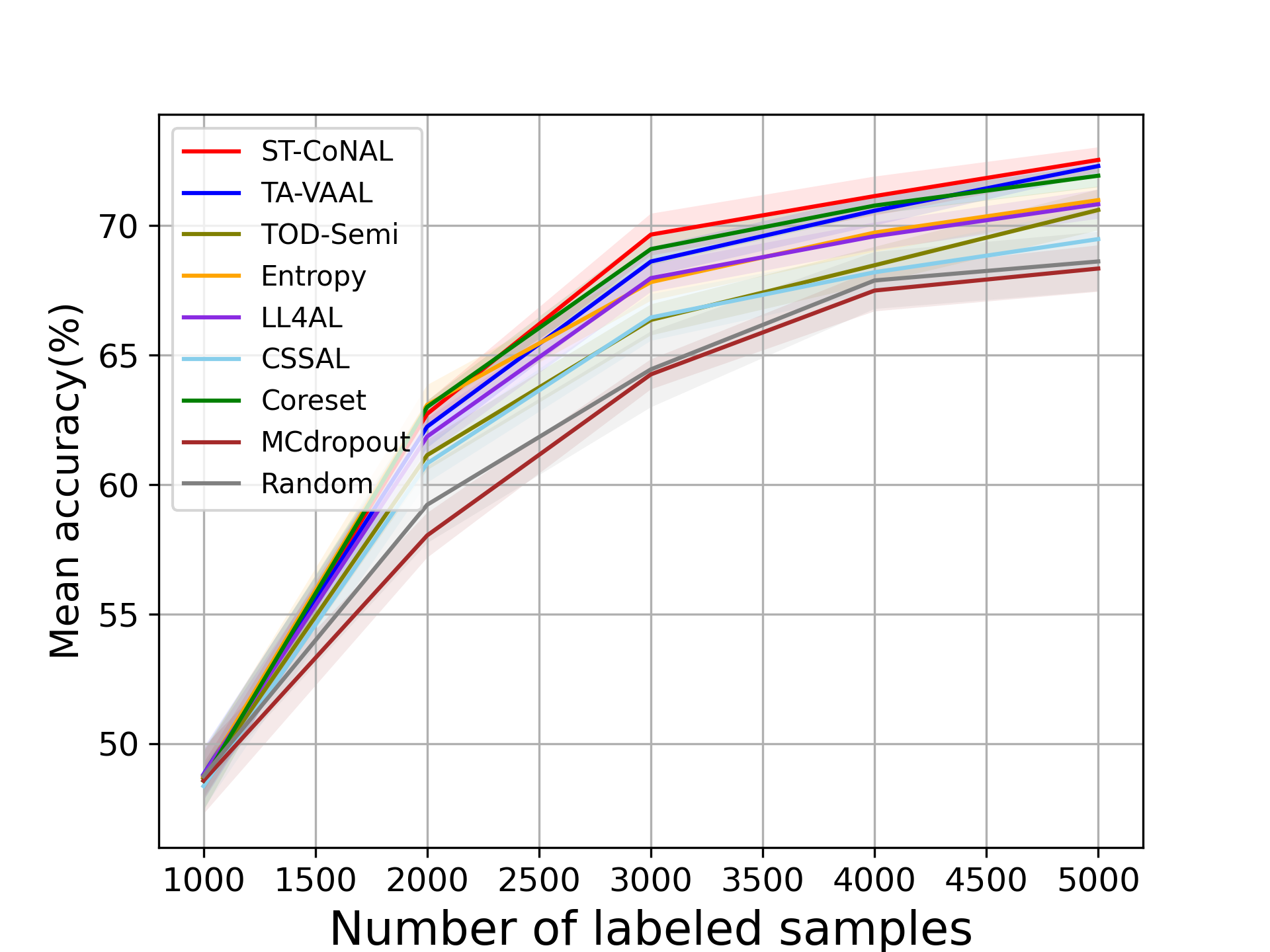}
        \caption[]{}
        \label{fig:caltech sl absolute}
    \end{subfigure}
    \hfill
    \begin{subfigure}[b]{0.490\textwidth}  
        \centering 
        \includegraphics[width=1.0\textwidth]{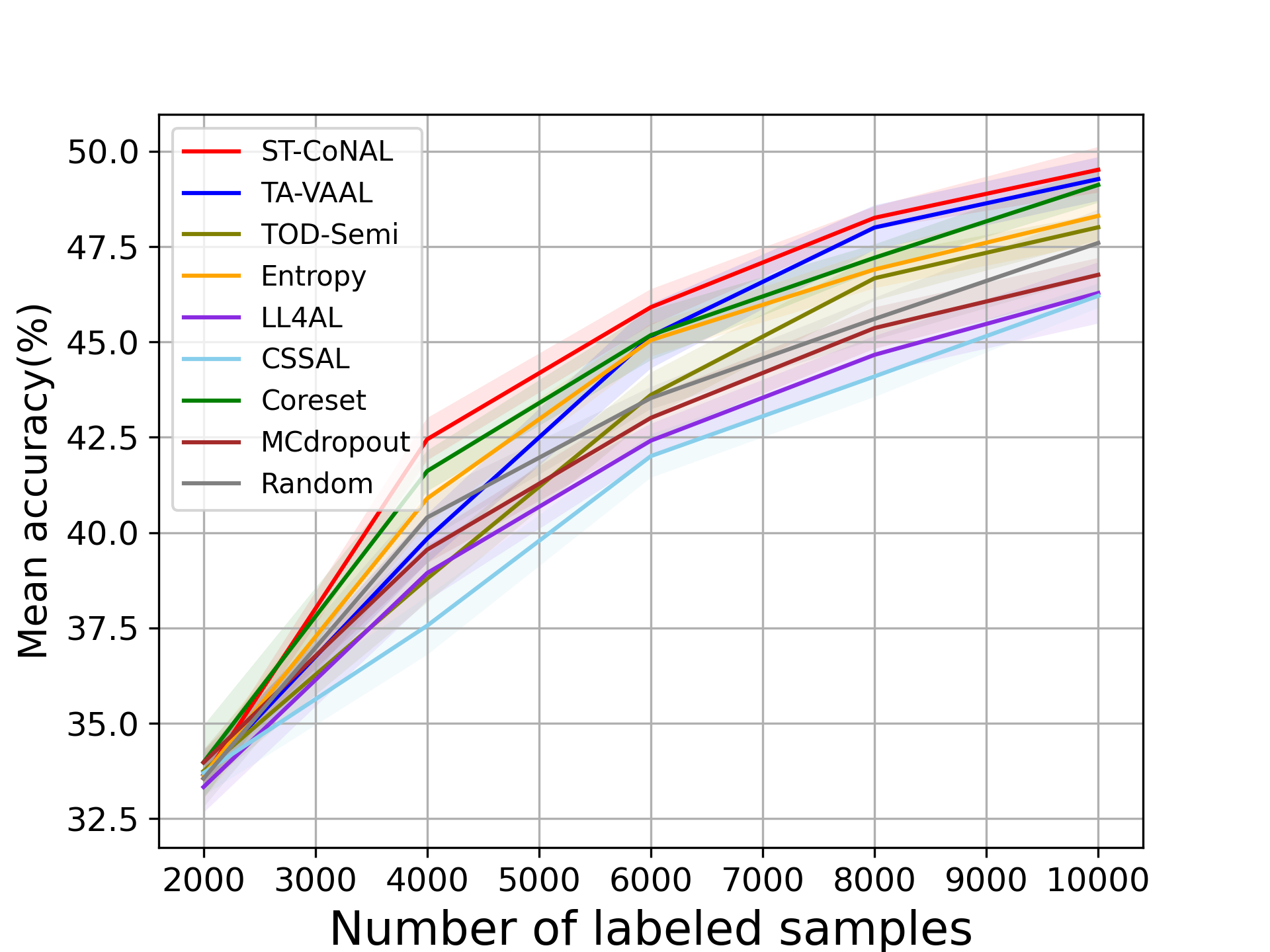}
        \caption[]{}
        \label{fig:tiny sl absolute}
    \end{subfigure}
    \captionsetup{singlelinecheck = false}
    \caption[]
    {Average test accuracy versus the number of labeled samples on (a) Caltech-256 and (b) Tiny ImageNet dataset.}
    \label{fig:caltech tiny sl absolute}
\end{figure*}

In the main manuscript, we used the performance gap from the random sampling baseline as a performance measure. To supplement this, we provide the absolute values of the average accuracy achieved by the AL methods. Fig. \ref{fig:cifar sl absolute} (a) and (b) show the performance results evaluated on CIFAR-10 and CIFAR-100 datasets and Fig. \ref{fig:caltech tiny sl absolute} (a) and (b) represent those on Caltech-256 and Tiny ImageNet datasets.

\section{Performance Comparison of AL Methods on Differently Imbalanced Datasets}
\begin{figure*}[t]
    % \centering
    \begin{subfigure}[b]{0.490\textwidth}
        \centering
        \includegraphics[width=1.0\textwidth]{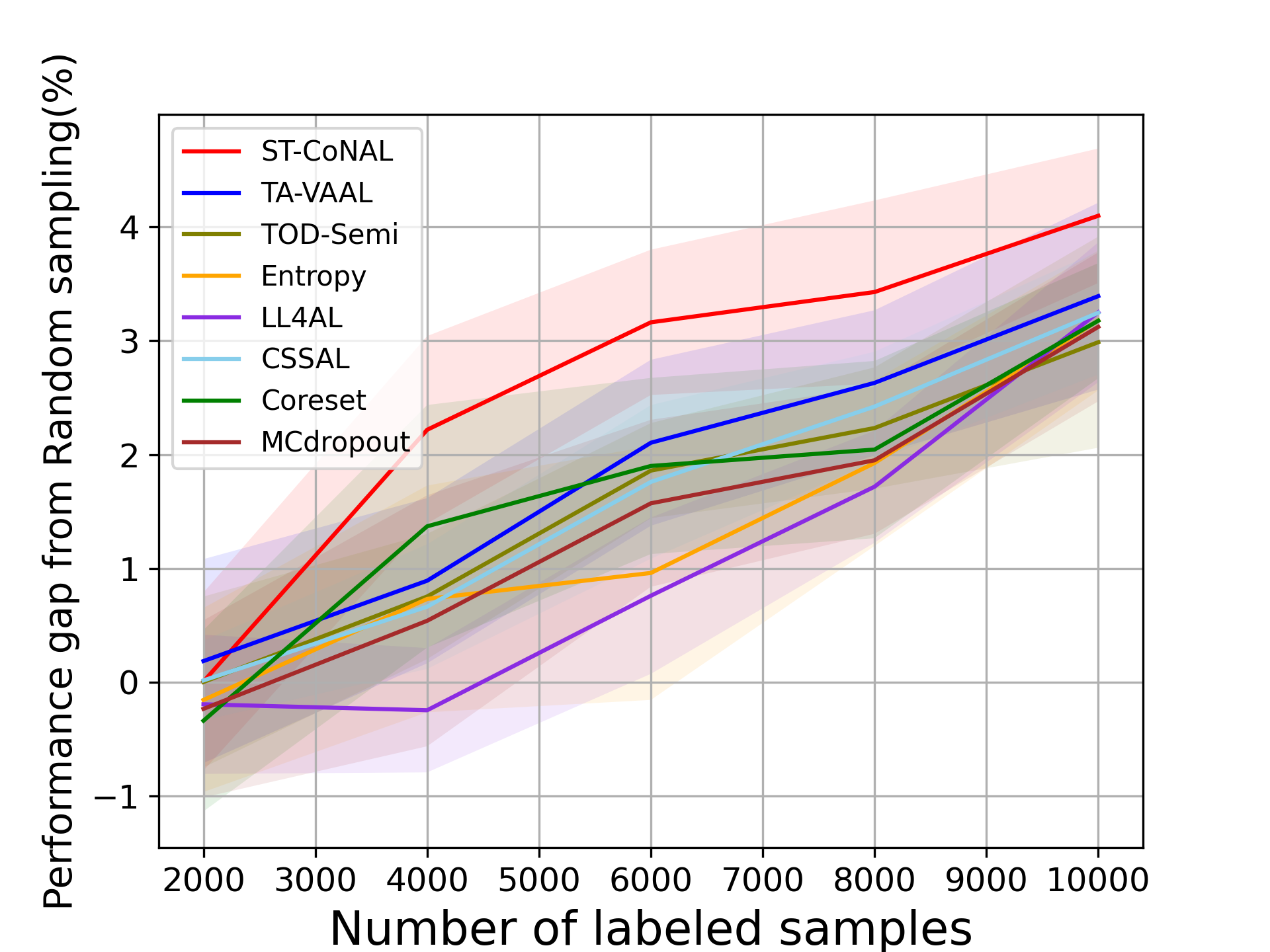}
        \caption[]{}
        \label{fig:cifar-100 im}
    \end{subfigure}
    \hfill
    \begin{subfigure}[b]{0.490\textwidth}  
        \centering 
        \includegraphics[width=1.0\textwidth]{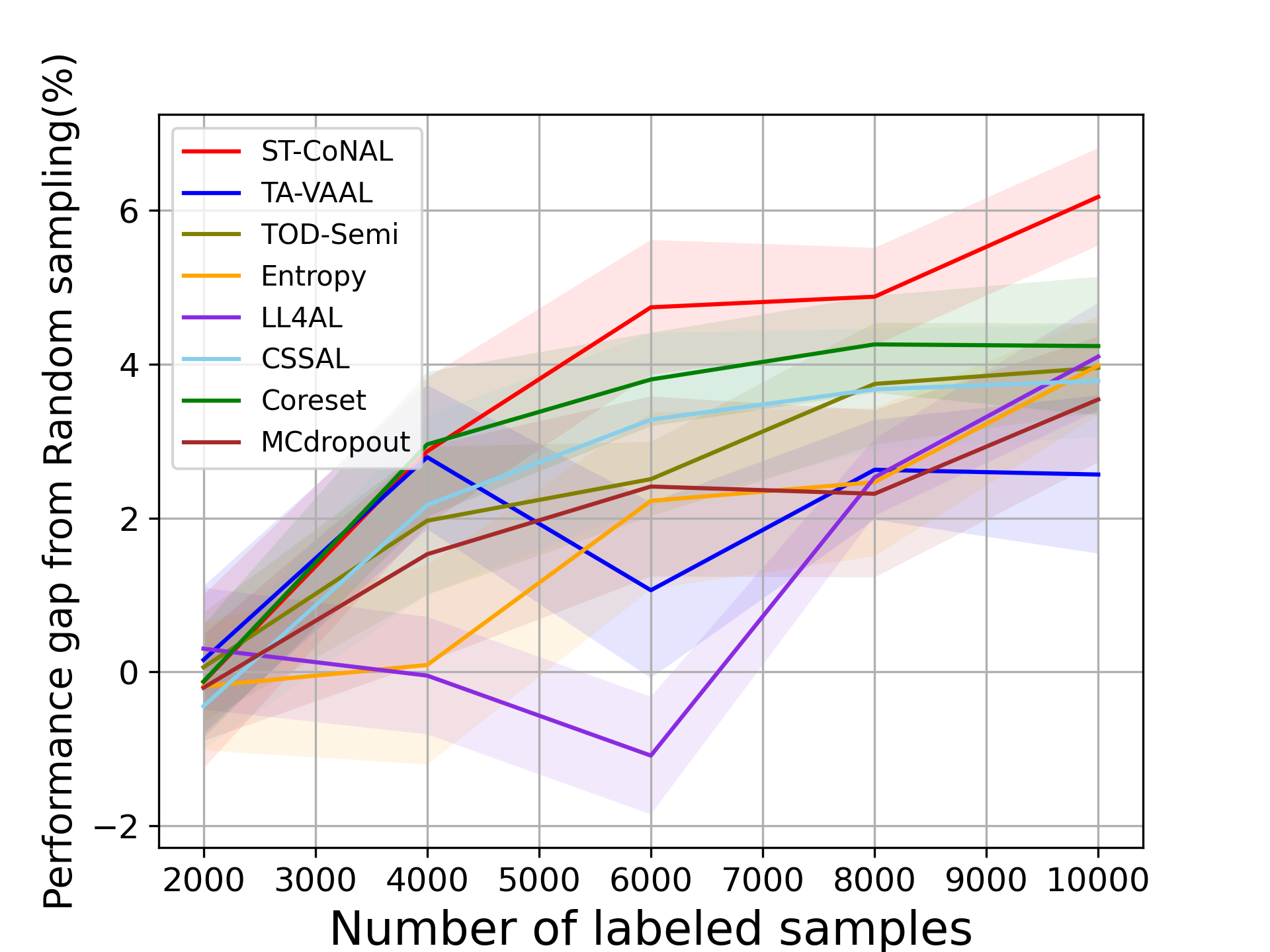}
        \caption[]{}
        \label{fig:cifar-100 imlong}
    \end{subfigure}
    \captionsetup{singlelinecheck = false}
    \caption[]
    {Average accuracy improvement from random sampling versus the number of labeled samples on (a) the step imbalanced CIFAR-100 and (b) the long-tailed CIFAR-100. The imbalance ratio was set to 100 for both cases.}
    \label{fig:cifar100 imbalance}
\end{figure*}

In the main manuscript, we reported the performance of ST-CoNAL obtained on two imbalanced versions of CIFAR-10. In this section,  we evaluate the algorithms on two differently imbalanced CIFAR-100 datasets, the step-imbalanced CIFAR-100 \cite{cui2019class} and long-tailed CIFAR-100 \cite{cao2019learning, cui2019class}. The imbalance ratio was set to 100 for all cases. Fig. \ref{fig:cifar100 imbalance} (a) and (b) present the performance of ST-CoNAL on the step imbalanced  CIFAR-100 \cite{kim2021task} and the long-tailed CIFAR-100 \cite{cao2019learning}, respectively. Even when different imbalance setups are used, ST-CoNAL consistently outperforms other AL methods. After the last acquisition step, ST-CoNAL achieves a performance improvement of 4.10\% and 6.18\% over the random sampling on the step-imbalanced CIFAR-100 and long-tailed CIFAR-100, respectively.

\section{Performance Comparison of AL Methods With Different Backbones}
\begin{figure*}[ht]
    % \centering
    \begin{subfigure}[b]{0.490\textwidth}
        \centering
        \includegraphics[width=1.0\textwidth]{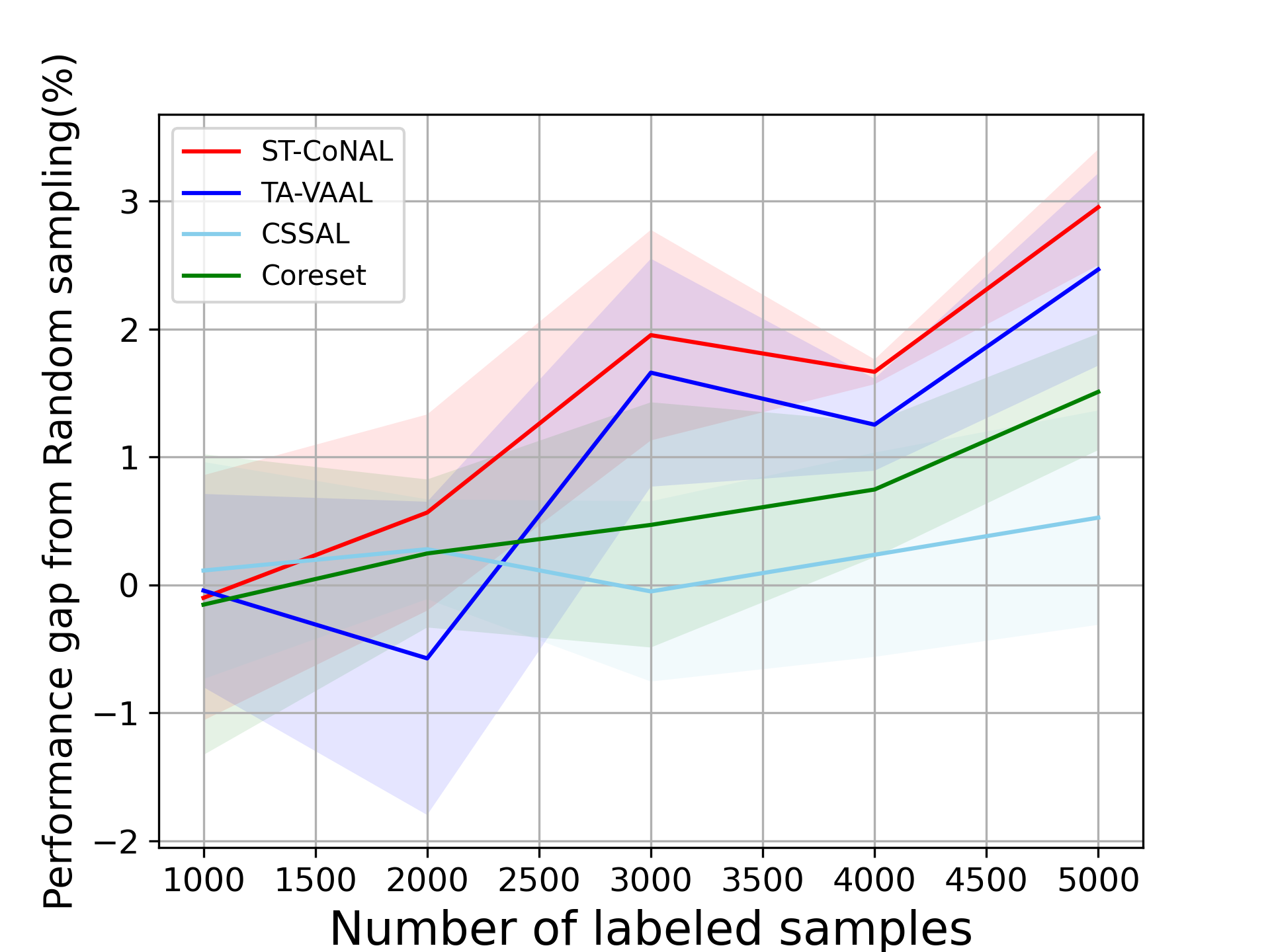}
        \caption[]{}
        \label{fig:cifar-10 vgg}
    \end{subfigure}
    \hfill
    \begin{subfigure}[b]{0.490\textwidth}  
        \centering 
        \includegraphics[width=1.0\textwidth]{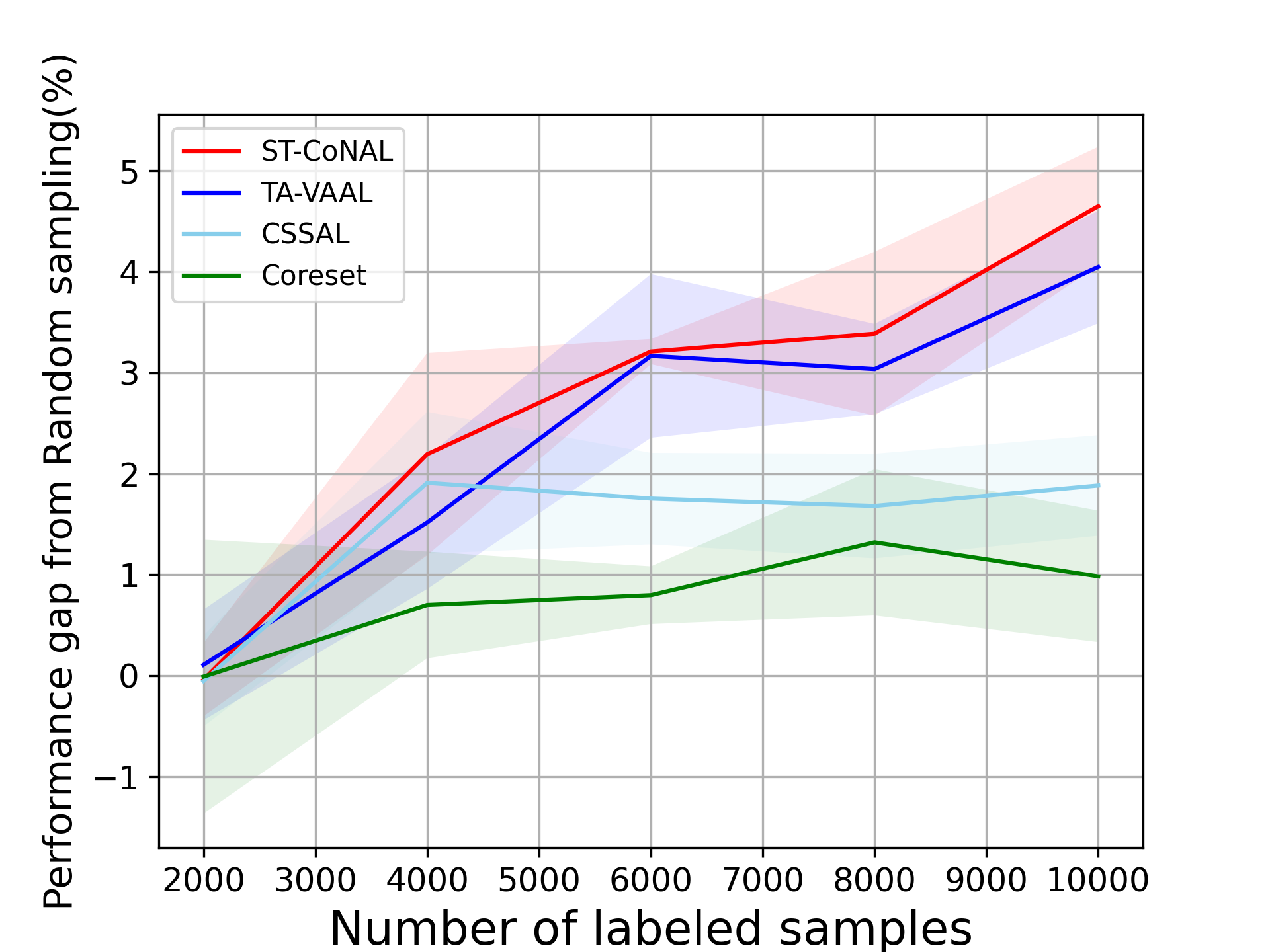}
        \caption[]{}
        \label{fig:cifar-100 vgg}
    \end{subfigure}
    \captionsetup{singlelinecheck = false}
    \caption[]
    {Average accuracy improvement from random sampling versus the number of labeled samples evaluated on CIFAR-10 and (b) CIFAR-100. We evaluate the performance of AL methods using VGG16 backbone \cite{simonyan2014very}.}
    \label{fig:cifar vgg}
\end{figure*}

\begin{figure*}[ht]
    \begin{subfigure}[]{0.49\textwidth}   
        \centering 
        \includegraphics[width=1.0\textwidth]{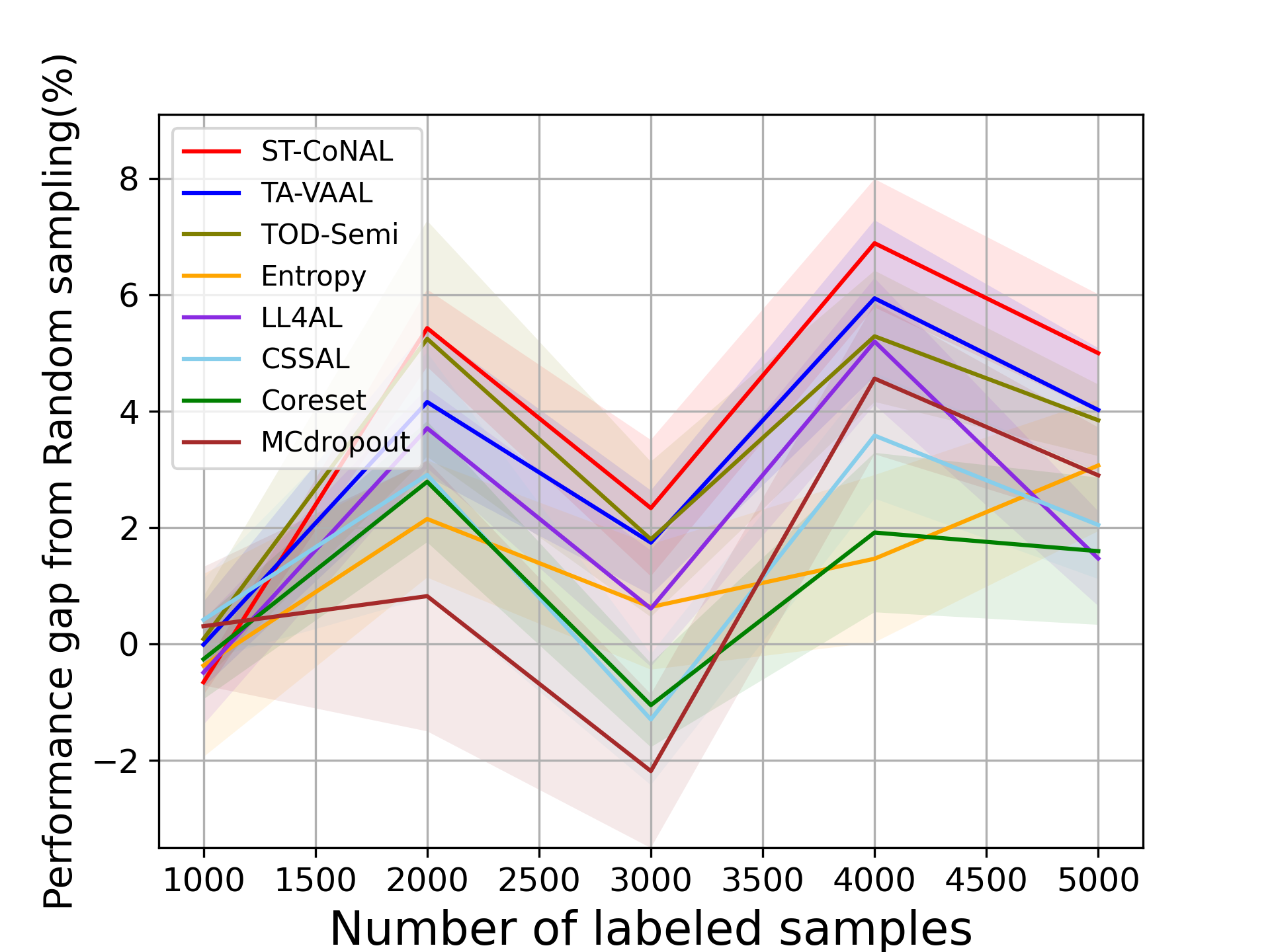} 
        \caption[]{}
        \label{fig:ResNet-50 on cifar-10}
    \end{subfigure}
    \hfill
    \begin{subfigure}[]{0.49\textwidth}   
        \centering 
        \includegraphics[width=1.0\textwidth]{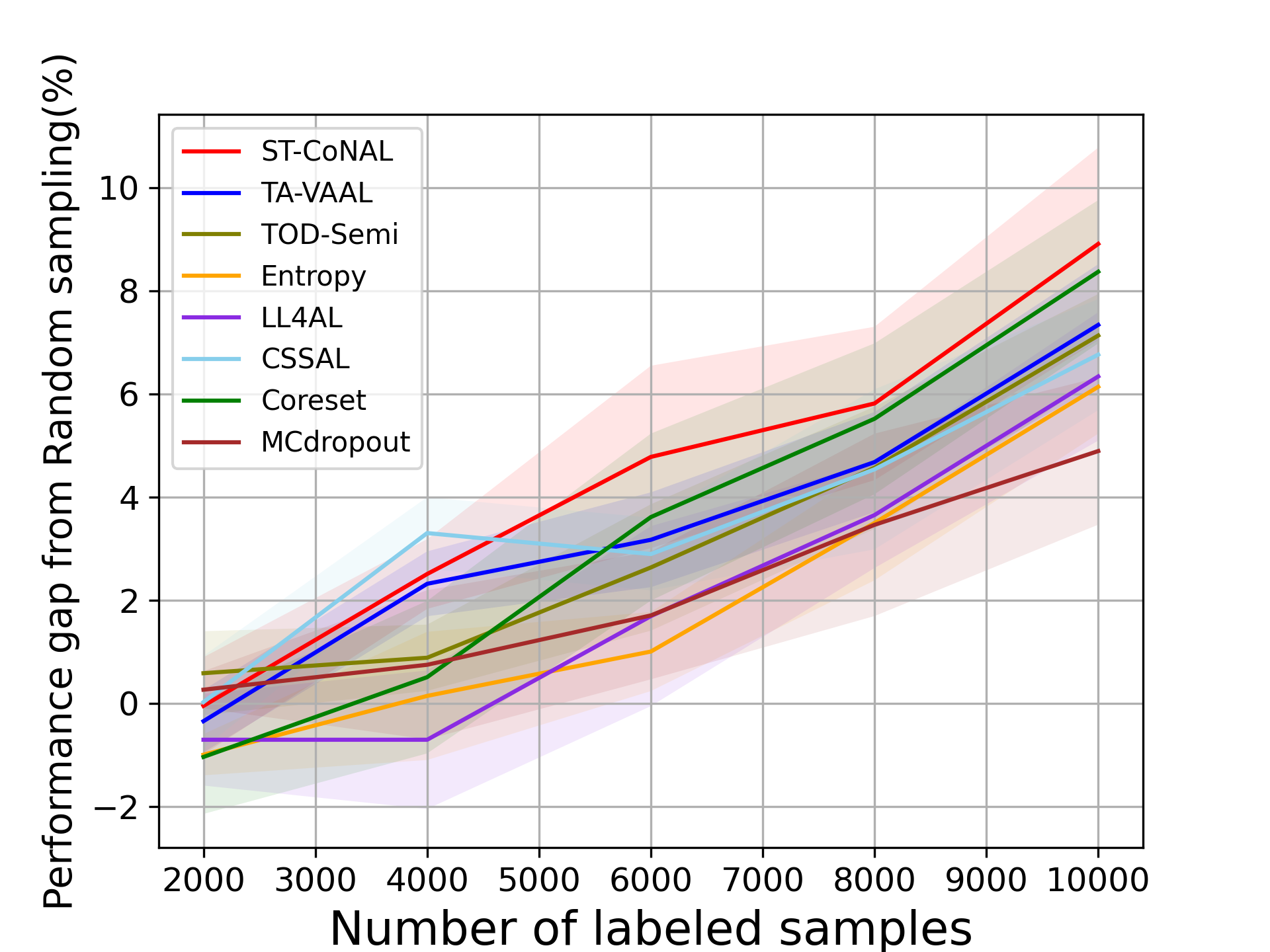} %, height=0.6\textwidth
        \caption[]{}
        \label{fig:ResNet-50 on cifar-100}
    \end{subfigure}
    \captionsetup{singlelinecheck = false}
    \caption[]
    {Average accuracy improvement from random sampling versus the number of labeled samples evaluated on (a) CIFAR-10 and (b) CIFAR-100. We evaluate the performance of AL methods using ResNet-50 backbone \cite{he2016deep}.}
    \label{fig:cifar resnet}
\end{figure*}

In this section, we evaluate the performance of ST-CoNAL when VGG16 \cite{simonyan2014very} and ResNet-50 \cite{he2016deep} are used as a backbone network.  Fig. \ref{fig:cifar vgg} (a) and (b) show the performance of ST-CoNAL  on  CIFAR-10 and CIFAR-100 when VGG16 backbone is used. After the last acquisition step, ST-CoNAL achieves 2.95\% and 4.65\% performance gains over random sampling on CIFAR-10 and  CIFAR-100, respectively. Fig. \ref{fig:cifar resnet} presents the performance of ST-CoNAL when ResNet-50 backbone is used. ST-CoNAL achieves the performance improvement of 4.99\% and 8.91\% over random sampling on CIFAR-10 and CIFAR-100, respectively. 

\section{Performance Versus Other Parameters}
We evaluate the performance of ST-CoNAL as a function of the budget sizes $b$, the number of student models $Q$ and temperature parameter $T$ on CIFAR-10. We try the different values of $b\in\{500, 1k, 2k\}$, $Q\in\{2, 4, 8, 10\}$, and $T\in\{0.3, 0.5, 0.7, 1.0\}$. For the budget sizes $b$, we compare our ST-CoNAL with three competitive methods, Entropy \cite{shannon1948mathematical}, and TA-VAAL \cite{gao2020consistency}. Table \ref{table:budget} provides the classification accuracy as a function of the acquisition step for different values of labeling budget $b$. The proposed ST-CoNAL maintains the performance gain over Entropy \cite{shannon1948mathematical} and TA-VAAL \cite{gao2020consistency} with different budget sizes. Additionally, we provide the performance of ST-CoNAL as a function of acquisition step versus the number of student models $Q$ and temperature parameter $T$. Table \ref{table:QandT} shows that ST-CoNAL achieves good performance for different $Q$ and $T$ values. For CIFAR-10 dataset, we set $Q=4$ and $T=0.7$ to provide decent performance.

\clearpage
\begin{table*}[h]
  \centering
  \begin{adjustbox}{width=0.9\columnwidth, center}
  \begin{tabular}{c|ccccccccccccccc}
  \hline\hline
  $b$ = $500$ & \multicolumn{3}{c}{$1k$ samples labeled} & \multicolumn{3}{c}{$2k$ samples labeled} &\multicolumn{3}{c}{$3k$ samples labeled} & \multicolumn{3}{c}{$4k$ samples labeled} & \multicolumn{3}{c}{$5k$ samples labeled}\\
  \hline
%   Random &   \multicolumn{3}{c}{46.95} & \multicolumn{3}{c}{58.19} & \multicolumn{3}{c}{68.18} & \multicolumn{3}{c}{74.93} & \multicolumn{3}{c}{78.24}\\
  Entropy &   \multicolumn{3}{c}{46.83} & \multicolumn{3}{c}{59.29} & \multicolumn{3}{c}{68.33} & \multicolumn{3}{c}{76.10} & \multicolumn{3}{c}{81.16}\\
  TA-VAAL &   \multicolumn{3}{c}{47.26} & \multicolumn{3}{c}{60.71} & \multicolumn{3}{c}{69.71} & \multicolumn{3}{c}{76.35} & \multicolumn{3}{c}{82.27}\\
  % in0-m1
  ST-CoNAL &   \multicolumn{3}{c}{46.93} & \multicolumn{3}{c}{\bf{60.91}} & \multicolumn{3}{c}{\bf{70.39}} & \multicolumn{3}{c}{\bf{77.03}} & \multicolumn{3}{c}{\bf{83.40}}\\
  \hline\hline
  $b$ = $1k$ & \multicolumn{3}{c}{$1k$ samples labeled} & \multicolumn{3}{c}{$2k$ samples labeled} &\multicolumn{3}{c}{$3k$ samples labeled} & \multicolumn{3}{c}{$4k$ samples labeled} & \multicolumn{3}{c}{$5k$ samples labeled}\\
  \hline
%   Random &   \multicolumn{3}{c}{47.04} & \multicolumn{3}{c}{58.80} & \multicolumn{3}{c}{67.86} & \multicolumn{3}{c}{75.20} & \multicolumn{3}{c}{78.60}\\
  Entropy &   \multicolumn{3}{c}{46.96} & \multicolumn{3}{c}{61.00} & \multicolumn{3}{c}{71.69} & \multicolumn{3}{c}{78.09} & \multicolumn{3}{c}{81.92}\\
  TA-VAAL &   \multicolumn{3}{c}{46.94} & \multicolumn{3}{c}{60.79} & \multicolumn{3}{c}{71.40} & \multicolumn{3}{c}{78.87} & \multicolumn{3}{c}{82.40}\\
  ST-CoNAL &   \multicolumn{3}{c}{47.13} & \multicolumn{3}{c}{\bf{61.48}} & \multicolumn{3}{c}{\bf{71.85}} & \multicolumn{3}{c}{\bf{79.61}} & \multicolumn{3}{c}{\bf{83.05}}\\
  \hline\hline
  $b$ = $2k$ & \multicolumn{3}{c}{$1k$ samples labeled} & \multicolumn{3}{c}{} & \multicolumn{3}{c}{$3k$ samples labeled} & \multicolumn{3}{c}{} & \multicolumn{3}{c}{$5k$ samples labeled}\\
  \hline
%   Random &   \multicolumn{3}{c}{46.91} & \multicolumn{3}{c}{} & \multicolumn{3}{c}{66.48} & \multicolumn{3}{c}{} & \multicolumn{3}{c}{77.12}\\
  Entropy &  \multicolumn{3}{c}{47.19} & \multicolumn{3}{c}{}& \multicolumn{3}{c}{71.00} & \multicolumn{3}{c}{} &  \multicolumn{3}{c}{81.29}\\
  TA-VAAL &  \multicolumn{3}{c}{47.18} & \multicolumn{3}{c}{}& \multicolumn{3}{c}{71.22} & \multicolumn{3}{c}{} &  \multicolumn{3}{c}{81.40}\\
  % in0-m1
  ST-CoNAL &  \multicolumn{3}{c}{46.99} & \multicolumn{3}{c}{}& \multicolumn{3}{c}{\bf{72.48}} & \multicolumn{3}{c}{} &  \multicolumn{3}{c}{\bf{82.53}}\\
  \hline
  \hline
  \end{tabular}
  \end{adjustbox}
  \caption{Mean accuracy versus the number of labeled data samples as the function of budget size $b$.}
  \label{table:budget}
\end{table*}
\begin{table*}[h]
  \centering
  \begin{adjustbox}{width=0.9\columnwidth, center}
  \begin{tabular}{c|ccccccccccccccc}
  \hline\hline
  Q & \multicolumn{3}{c}{$1k$ samples labeled} & \multicolumn{3}{c}{$2k$ samples labeled} &\multicolumn{3}{c}{$3k$ samples labeled} & \multicolumn{3}{c}{$4k$ samples labeled} & \multicolumn{3}{c}{$5k$ samples labeled}\\
  \hline
  2 & \multicolumn{3}{c}{46.67} & \multicolumn{3}{c}{60.99} & \multicolumn{3}{c}{70.95} & \multicolumn{3}{c}{79.18} & \multicolumn{3}{c}{82.63}\\
  4 & \multicolumn{3}{c}{47.13} & \multicolumn{3}{c}{61.48} & \multicolumn{3}{c}{\bf{71.85}} & \multicolumn{3}{c}{\bf{79.61}} & \multicolumn{3}{c}{83.05}\\
  8 & \multicolumn{3}{c}{46.86} & \multicolumn{3}{c}{\bf{61.49}} & \multicolumn{3}{c}{71.77} & \multicolumn{3}{c}{79.23} & \multicolumn{3}{c}{82.83}\\
  10 & \multicolumn{3}{c}{46.70} & \multicolumn{3}{c}{61.37} & \multicolumn{3}{c}{71.64} & \multicolumn{3}{c}{79.40} & \multicolumn{3}{c}{\bf{83.14}}\\
  \hline\hline
  T & \multicolumn{3}{c}{$1k$ samples labeled} & \multicolumn{3}{c}{$2k$ samples labeled} &\multicolumn{3}{c}{$3k$ samples labeled} & \multicolumn{3}{c}{$4k$ samples labeled} & \multicolumn{3}{c}{$5k$ samples labeled}\\
  \hline
   0.3 & \multicolumn{3}{c}{47.00} & \multicolumn{3}{c}{60.66} & \multicolumn{3}{c}{70.62} & \multicolumn{3}{c}{78.03} & \multicolumn{3}{c}{82.57}\\
   0.5 & \multicolumn{3}{c}{47.23} & \multicolumn{3}{c}{60.91} & \multicolumn{3}{c}{70.82} & \multicolumn{3}{c}{78.51} & \multicolumn{3}{c}{82.35}\\
   0.7 & \multicolumn{3}{c}{47.13} & \multicolumn{3}{c}{\bf{61.48}} & \multicolumn{3}{c}{\bf{71.85}} & \multicolumn{3}{c}{79.61} & \multicolumn{3}{c}{\bf{83.05}}\\
   1.0 & \multicolumn{3}{c}{46.87} & \multicolumn{3}{c}{60.91} & \multicolumn{3}{c}{71.41} & \multicolumn{3}{c}{\bf{79.78}} & \multicolumn{3}{c}{82.96}\\
  \hline\hline
  \end{tabular}
  \end{adjustbox}
  \caption{Mean accuracy versus the number of labeled data samples as the function of the number of student models $Q$ and temperature parameter $T$.}
  \label{table:QandT}
\end{table*}





% LNCS Homepage, \url{http://www.springer.com/lncs}. Last accessed 4
% \end{thebibliography}
\clearpage

\end{document}